\documentclass[runningheads]{llncs}
\usepackage{graphicx}
\usepackage{multirow}
\usepackage{soul}
\usepackage{subfigure}

\usepackage{tikz}
\usepackage{comment}
\usepackage{amsmath,amssymb} 
\usepackage{color}

\usepackage[accsupp]{axessibility}  
\usepackage[misc]{ifsym}

\newcommand{\Figref}[1]{Figure~\ref{fig:#1}}
\newcommand{\figref}[1]{Fig.~\ref{fig:#1}}
\newcommand{\Eqnref}[1]{Equation~\ref{eq:#1}}
\newcommand{\eqnref}[1]{Eq.~\ref{eq:#1}}
\newcommand{\Tabref}[1]{Table~\ref{tab:#1}}

\graphicspath{{./figures}{./}}

\begin{document}
\pagestyle{headings}
\mainmatter
\def\ECCVSubNumber{609}  

\title{Human Trajectory Prediction via Neural Social Physics} 

\titlerunning{Human Trajectory Prediction via Neural Social Physics}
%
\author{Jiangbei Yue\inst{1} \and
Dinesh Manocha\inst{2} \and
He Wang\inst{1}$^{(\textrm{\Letter})}$}
\authorrunning{J. Yue et al.}
%
\institute{University of Leeds, Leeds, UK \\ \email{H.E.Wang@leeds.ac.uk} \and
University of Maryland at College Park, College Park, USA}


\maketitle

\begin{abstract}
Trajectory prediction has been widely pursued in many fields, and many \textit{model-based} and \textit{model-free} methods have been explored. The former include rule-based, geometric or optimization-based models, and the latter are mainly comprised of deep learning approaches. In this paper, we propose a new method combining both methodologies based on a new Neural Differential Equation model. Our new model (Neural Social Physics or NSP) is a deep neural network within which we use an explicit physics model with learnable parameters. The explicit physics model serves as a strong inductive bias in modeling pedestrian behaviors, while the rest of the network provides a strong data-fitting capability in terms of system parameter estimation and dynamics stochasticity modeling. We compare NSP with 15 recent deep learning methods on 6 datasets and improve the state-of-the-art performance by 5.56\%-70\%. Besides, we show that NSP has better generalizability in predicting plausible trajectories in drastically different scenarios where the density is 2-5 times as high as the testing data. Finally, we show that the physics model in NSP can provide plausible explanations for pedestrian behaviors, as opposed to black-box deep learning. Code is available: https://github.com/realcrane/Human-Trajectory-Prediction-via-Neural-Social-Physics.
\keywords{Human Trajectory Prediction; Neural Differential Equations}
\end{abstract}

\section{Introduction}
Understanding human trajectories is key to many research areas such as physics, computer science and social sciences. 
Being able to learn behaviors with non-invasive sensors is important to analyzing the natural behaviors of humans. This problem has been widely studied in computer graphics, computer vision and machine learning~\cite{BENDALIBRAHAM_recent_2021}. Existing approaches generally fall into \textit{model-based} and \textit{model-free} methods. Early model-based methods tended to be empirical or rule-based methods derived via the first-principles approach: summarizing observations into rules and deterministic systems based on fundamental assumptions on human motion. In such a perspective, social interactions can be modelled as forces in a particle system~\cite{helbing1995social} or an optimization problem~\cite{vanDenBerg_reciprocal_2008}, and individuals can be influenced by affective states~\cite{Luo_Agentbased_2008}. Later, data-driven model-based methods were introduced, in which the model behavior is still dominated by the assumptions on the dynamics, e.g. a linear dynamical system~\cite{He_Informative_2020}, but retains sufficient flexibility so that the model can be adjusted to fit observations. 
More recently, model-free methods based on deep learning have also been explored, and these demonstrate surprising trajectory prediction capability~\cite{alahi2016social,gupta2018social,sadeghian2019sophie,bhattacharyya2019conditional,li2019conditional,liang2019peeking,deo2020trajectory,liang2020simaug,mangalam2020not,salzmann2020trajectron++,liang2020garden,mangalam2021goals,su2021pedestrian,zhou2021sliding,gao2022social,xia2022cscnet}.

Empirical or rule-based methods possess good explainability because they are formed as explicit geometric optimization or ordinary/partial differentiable equations where specific terms correspond to certain behaviors.
Therefore, they have been used for not only prediction but also analysis and simulation~\cite{vantoll_algorithms_2021}. However, they are less effective in data fitting with respect to noise and are therefore unable to predict accurately, even when the model is calibrated on data~\cite{Wolinski_paramter_2014}. Data-driven model-based methods (e.g., statistical machine learning) improve the ability of data fitting but are restricted by the specific statistical models employed which have limited capacities to learn from large amounts of data~\cite{He_Informative_2020}. Finally, deep learning approaches excel at data fitting. They can learn from large datasets, but lack explainability and therefore have been mainly used for prediction rather than analysis and simulation~\cite{alahi2016social,mangalam2020not,zhou2021sliding}.

We explore a model that can explain pedestrian behaviors and retain good data-fitting capabilities by combining model-based and model-free approaches. Inspired by recent research in neural differential equations~\cite{chen2018neural,rackauckas2020universal,zhong2019symplectic,zubov2021NeuralPDE,kidger_neural_2022}, we propose a new crowd neural differentiable equation model consisting of two parts. The first is a deterministic model formulated using a differentiable equation. Although this equation can be arbitrary, we use a dynamical system inspired by the social force model~\cite{helbing1995social}. In contrast to the social force model and its variants, the key parameters of our deterministic model are learnable through data instead of being hand-picked and fixed. The second part of our model captures complex uncertainty in the motion dynamics and observations via a Variational Autoencoder. Overall, the whole model is a deep neural network with an embedded explicit model; we call this model \textit{Neural Social Physics} (NSP).

We demonstrate that our NSP model outperforms the state-of-the-art methods~\cite{gupta2018social,sadeghian2019sophie,bhattacharyya2019conditional,li2019conditional,liang2019peeking,deo2020trajectory,liang2020simaug,mangalam2020not,salzmann2020trajectron++,liang2020garden,mangalam2021goals,su2021pedestrian,zhou2021sliding,gao2022social,xia2022cscnet} in standard trajectory prediction tasks across various benchmark datasets~\cite{robicquet2016learning,pellegrini2010improving,lerner2007crowds} and metrics. In addition, we show that NSP can generalize to unseen scenarios with higher densities and still predict plausible motions with less collision between people, as opposed to pure black-box deep learning approaches. Finally, from the explicit model in NSP, we demonstrate that our method can provide plausible explanations for motions. 
Formally, (1) we propose a new neural differentiable equation model for trajectory prediction and analysis. (2) we propose a new mechanism to combine explicit and deterministic models with deep neural networks for crowd modeling. (3) We demonstrate the advantages of the NSP model in several aspects: prediction accuracy, generalization and explaining behaviors.

\section{Related Work}
\subsection{Trajectory Analysis and Prediction}
Statistical machine learning has been used for trajectory analysis in computer vision~\cite{oliver2000bayesian,ellis2009modelling,wang2011trajectory,kim2015interactive,wang2016globally,chaker2017social}. They aim to learn individual motion dynamics~\cite{zhou2011random}, structured latent patterns in data~\cite{wang2016globally,wang_trending_2016}, anomalies~\cite{charalambous2014data,chaker2017social}, etc. These methods provide a certain level of explainability, but are limited in model capacity for learning from large amounts of data. Compared
with these methods, our model leverages the ability of deep neural networks to handle high-dimensional and large data. More recently, deep learning has been exploited for trajectory prediction~\cite{sighencea2021review}. Recurrent neural networks (RNNs)~\cite{alahi2016social,bartoli2018context,vemula2018social} have been explored first due to their ability to learn from temporal data. Subsequently, other  deep learning techniques and neural network architectures are introduced into trajectory prediction, such as Generative Adversarial Network (GAN)~\cite{gupta2018social}, conditional variational autoencoder (CVAE)~\cite{ivanovic2019trajectron,mangalam2020not,zhou2021sliding} and Convolutional Neural Network (CNN)~\cite{mohamed2020social}. In order to capture the spatial features of trajectories and the interactions between pedestrians accurately, graph neural networks (GNNs) have also been used to reason and predict future trajectories~\cite{mohamed2020social,shi2021sgcn}. 
Compared with existing deep learning methods, our method achieves better prediction accuracy. Further, our method has an explicit model which can explain pedestrian motions and lead to better generalizability. Very recently, attempts have been made in combining physics with deep learning for trajectory prediction~{\cite{antonucci2020generating,kreiss2021deep,hossain2022sfmgnet}}. But their methods are tied to specific physics models and are deterministic, while NSP is a general framework that aims to accommodate arbitrary physics models and is designed to be intrinsically stochastic to capture motion randomness.

\subsection{Pedestrian and Crowd Simulation}
Crowd simulation aims to generate trajectories given the initial position and destination of each agent~\cite{vantoll_algorithms_2021}, which essentially aims to predict individual motions. Empirical modelling and data-driven methods have been the two foundations in simulation~\cite{narain2009aggregate,lopez2019character}. Early research is dominated by empirical modelling or rule-based methods, where crowd motions are abstracted into mathematical equations and deterministic systems, such as flows~\cite{narain2009aggregate}, particle systems~\cite{helbing1995social}, and velocity and geometric optimization~\cite{vanDenBerg_reciprocal_2008,shen2018data}. Meanwhile, data-driven methods using statistical machine learning have also been employed, e.g., using first-person vision to guide steering behaviors~\cite{lopez2019character} or using trajectories to extract features to describe motions~\cite{karamouzas2018crowd,wei2020simulating}. While the key parameters in these approaches are either fixed or learned from small datasets, our NSP model is more general. It can take existing deterministic systems as a component and provides better data-fitting capacity via deep neural networks. Compared with afore-mentioned model-based methods, our NSP can be regarded as using deep learning for model calibration. our model possesses the ability to learn from large amount of data, which is difficult for traditional parameter estimation methods based on optimization or sampling~{\cite{wan2017learning}}. Meanwhile, the formulation of our NSP is more general, flexible and data-driven than traditional model-based methods.

\subsection{Deep Learning and Differential Equations}
Solving differentiable equations (DE) with the assistance of deep learning has recently spiked strong interests~\cite{chen2018neural,zhong2019symplectic,zubov2021NeuralPDE,Karniadakis_datadriven_2021}. Based on the involvement depth of deep learning, the research can be categorized into deep learning assisted DE, differentiable physics, neural differential equations and physics-informed neural networks (PINNs). Deep learning assisted DE involves accelerating various steps during the DE solve, such as Finite Element mesh generation~\cite{Zhang_MeshingNet_2020,zhang_meshingnet3d_2021}. The deeper involvement of neural networks is shown in differentiable physics and neural differential equations, where the former aims to make the whole simulation process differentiable~\cite{Gong_finegrained_2022,liang2019differentiable,Werling2021fast}, and the latter focuses on the part of the equations being parameterized by neural networks~\cite{Shen_high_2021}. PINNs aim to bypass the DE solve and use NN for prediction~\cite{raissi2019physics,cai2022physics}. Highly inspired by the research above, we propose a new neural differential equations model in a new application domain for human trajectory prediction.

\section{Methodology}

\subsection{Neural Social Physics (NSP)}

At any time $t$, the position $p^t_i$ of the $i$th pedestrian can be observed in a crowd. Then a trajectory can be represented as a function of time $q(t)$, where we have discrete observations in time up to $T$, $\{q^0, q^1, \cdots, q^T\}$. An observation or \textit{state} of a person at time $t$ is represented by $q^t = [p^t, \dot{p}^t]^\mathbf{T}$ where $p$, $\dot{p} \in \mathbb{R}^2$ are the position and velocity. For most datasets, $p$ is given and $\dot{p}$ can be estimated via finite difference. Given an observation $q_n^t$ of the $n$th person, we consider her neighborhood set $\Omega_n^t$ containing other nearby pedestrians $\{q_j^t: j \in \Omega_n^t\}$. The neighborhood is also a function of time $\Omega(t)$. Then, in NSP the dynamics of a person (agent) in a crowd can be formulated as:
\begin{equation}
\label{eq:nde}
    \frac{dq}{dt}(t) = f_{\theta, \phi}(t, q(t), \Omega(t), q^T, E) + \alpha_{\phi}(t, q^{t:t-M})
\end{equation}
where $\theta$ and $\phi$ are learnable parameters, $E$ represents the environment. $\theta$ contains interpretable parameters explained later and $\phi$ contains uninterpretable parameters (e.g. neural network weights). The agent dynamics are governed by $f$ which depends on time $t$, its current state $q(t)$, its time-varying neighborhood $\Omega(t)$ and the environment $E$. Similar to existing work, we assume there is dynamics stochasticity in NSP. But unlike them which assume simple forms (e.g. white noise)~\cite{He_Informative_2020}, we model time-varying stochasticity in a more general form: as a function of time, the current state and the brief history of the agent, $\alpha_{\phi}(t, q^{t:t-M})$. Then we have the following equation in NSP:
\begin{equation}
    q^T = q^0 + \int_{t=0}^T {f_{\theta, \phi}(t, q(t), \Omega(t), q^T, E) + \alpha_{\phi}(t, q^{t:t-M})} dt
\end{equation}
given the initial and final condition $q(0) = q^0 \text{ and } q(T) = q^T$.

Physics models have been widely used to model crowd dynamics~\cite{narain2009aggregate,helbing1995social}. To leverage their interpretability, we model the dynamics as a physical system in NSP. Assuming the second-order differentiability of $p(t)$, NSP expands $q(t)$ via Taylor's series for a first-order approximation:
\begin{equation}
\label{eq:timeDiscretization}
    q(t+\triangle t) \approx q(t) + \dot{q}(t) \triangle t =
    \begin{pmatrix}
    p(t)\\
    \dot{p}(t)
    \end{pmatrix}
    + \triangle t
    \begin{pmatrix}
    \dot{p}(t) + \alpha(t, q^{t:t-M})\\
    \ddot{p}(t)
    \end{pmatrix}
\end{equation}
where $\triangle t$ is the time step. The stochasticity $\alpha(t, q^{t:t-M})$ is assumed to only influence $\dot{p}$. \Eqnref{timeDiscretization} is general and any dynamical system with second-order differentiability can be employed here. Below, we realize NSP by combining a type of physics models-social force models (SFM)~\cite{helbing1995social} and neural networks. We refer to our model NSP-SFM.

\subsection{NSP-SFM}
\begin{figure}[b]
\centering
\includegraphics[width=0.7\textwidth]{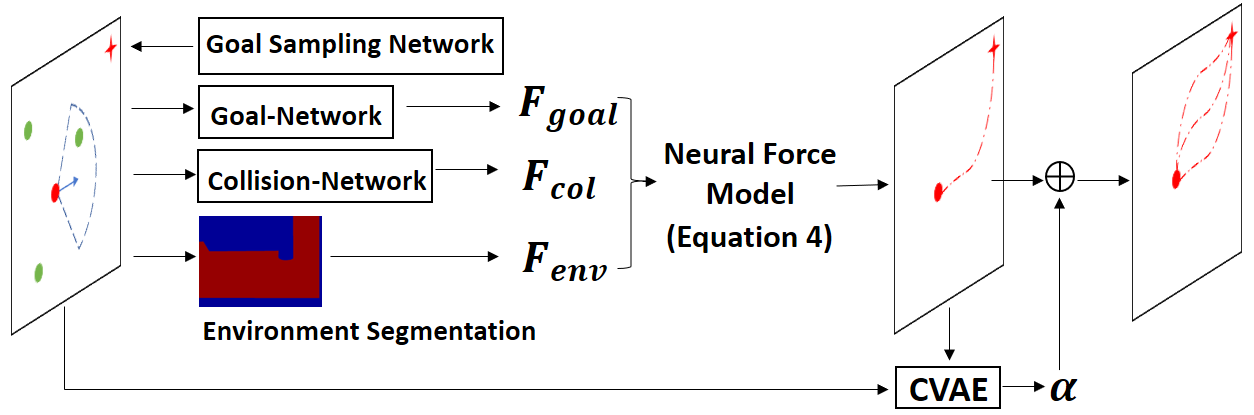}
\caption{Overview of NSP-SFM. $F_{goal}$, $F_{col}$ and $F_{env}$ are estimated in every time step by Goal-Network, Collision-Network and \eqnref{env} before solving \eqnref{forces}. The output is used to update the position and velocity which are then combined with the estimated noise from $\alpha$ for the final prediction}
\label{fig:model}
\end{figure}

We design the NSP-SFM by assuming each person acts as a particle in a particle system and each particle is governed by Newton's second law of motion. $\ddot{p}(t)$ is designed to be dependent on three forces: goal attraction $F_{goal}$, inter-agent repulsion $F_{col}$ and environment repulsion $F_{env}$.
\begin{align}
\label{eq:forces}
    \ddot{p}(t) = F_{goal}(t, q^T, q^t) + F_{col} (t, q^t, \Omega^t) + F_{env}(t, q^t, E)
\end{align}
where $E$ is the environment and explained later. However, unlike \cite{helbing1995social}, the three forces are partially realized by neural networks, turning \Eqnref{nde} into a neural differential equation.
The overall model is shown in~\Figref{model}. Note that, in \Eqnref{nde}, we assume $p^T$ is given, although it is not available during prediction. Therefore, we employ a Goal Sampling Network (GSN) to sample $p^T$. During testing, we either first sample a $p^T$ for prediction or require the user to input $p^T$. The GSN is similar to a part of Y-net~\cite{mangalam2021goals} and pre-trained, and detailed in the supplementary materials.

\begin{figure}[tb]
\centering
\includegraphics[width=0.8\textwidth]{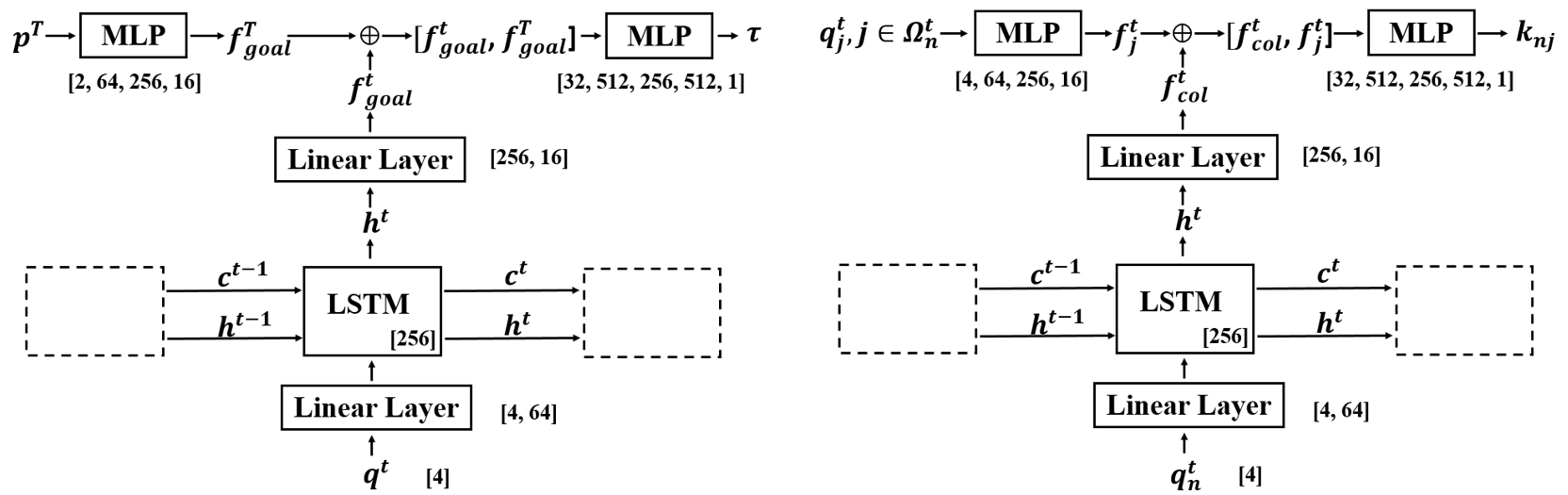}
\caption{Left: Goal-Network and Right: Collision-Network. The numbers in square brackets show both the number and dimension of the layers in each component.}
\label{fig:subnets}
\end{figure}

\begin{figure}[tb]
\centering
\includegraphics[width=0.8\textwidth]{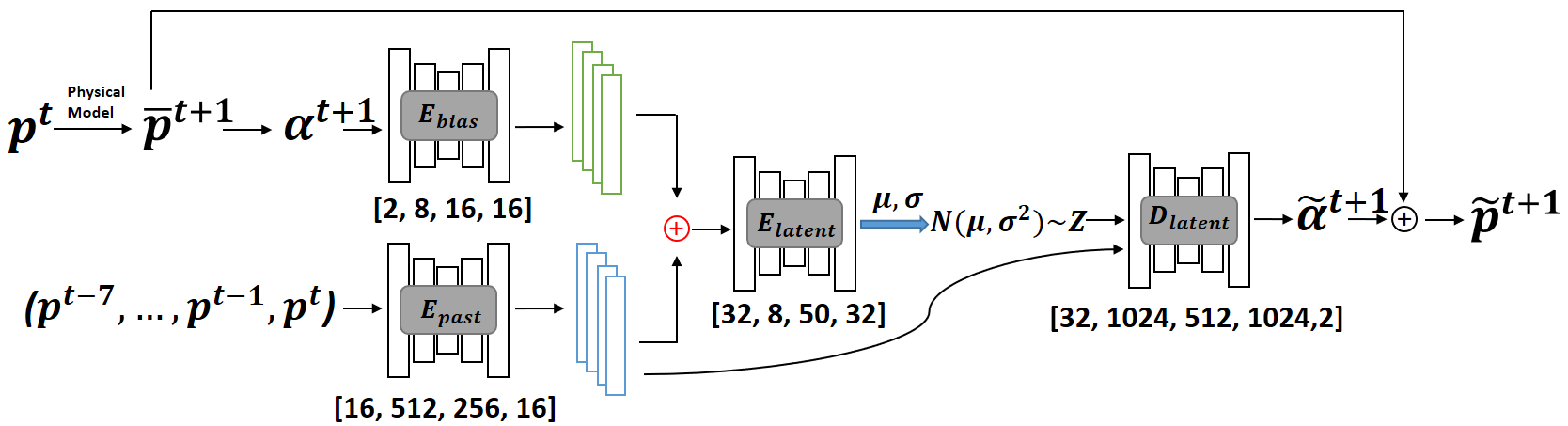}
\caption{The architecture of the CVAE, where $\Bar{p}^{t+1}$ is the intermediate prediction out of our force model and $\alpha^{t+1} = p^{t+1} - \Bar{p}^{t+1}$. Encoder $E_{bias}$, $E_{past}$, $E_{latent}$ and decoder $D_{latent}$ are all MLP networks with dimensions indicated in the square brackets. More Details of the network can be found in the supplementary material}
\label{fig:CVAE}
\end{figure}

Given the current state and the goal, we compute $F_{goal}$ using the Goal-Network $NN_{\phi_1}$ in \eqnref{goal} (\figref{subnets} Left), $F_{col}$ using the Collision-Network $NN_{\phi_2}$ in \eqnref{collision} (\figref{subnets} Right) and $F_{env}$ using \eqnref{env} directly. The Goal-Network encodes $q^t$ then feeds it into a Long Short Term Memory (LSTM) network to capture dynamics. After a linear transformation, the LSTM output is concatenated with the embedded $p^T$. Finally, $\tau$ is computed by an MLP (multi-layer perceptron). In Collision-Network, the architecture is similar. Every agent $q_j^t$ in the neighborhood $\Omega_n^t$ is encoded and concatenated with the encoded agent $q_n^t$. Then $k_{nj}$ is computed. $\tau$ and $k_{nj}$ are interpretable key parameters of $F_{goal}$ and $F_{col}$. The corresponding parameter in $F_{env}$ is $k_{env}$. Finally, we show our network for $\alpha$ for stochasticity modeling in \Figref{CVAE}.

\textbf{Goal attraction}. Pedestrians are always drawn to destinations, which can be abstracted into a goal attraction force. At time $t$, a pedestrian has a desired walking direction $e^t$ determined by the goal $p^T$ and the current position $p^t$: $e^t = \frac{p^T - p^t}{\lVert p^T - p^t \rVert}$. If there are no other forces, she will change her current velocity to the desired velocity $v_{des}^t = v_0^t e^t$ where $v_0^t$ and $e^t$ are the magnitude and direction respectively. Instead of using a fixed $v_0$ as in \cite{helbing1995social}, we update $v_0^t$ at every $t$ to mimic the change of the desired speed as the pedestrian approaches the destination: $v_0^t = \frac{\lVert p^T - p^t \rVert}{(T-t)\triangle t}$. Therefore, the desired velocity is defined as $v_{des}^t = v_0^t e^t = \frac{p^T - p^t}{(T-t)\triangle t}$.
The goal attraction force $F_{goal}$ represents the tendency of a pedestrian changing her current velocity $\dot{p}^t$ to the desired velocity $v_{des}^t$ within time $\tau$:
\begin{equation}
\label{eq:goal}
    F_{goal} = \frac{1}{\tau}(v_{des}^t - \dot{p}^t) \text { where } \tau = NN_{\phi_1}(q^t, p^T)
\end{equation}
where $\tau$ is learned through a neural network (NN) parameterized by $\phi_1$.

\begin{figure}[tb]
\centering
\includegraphics[width=\textwidth]{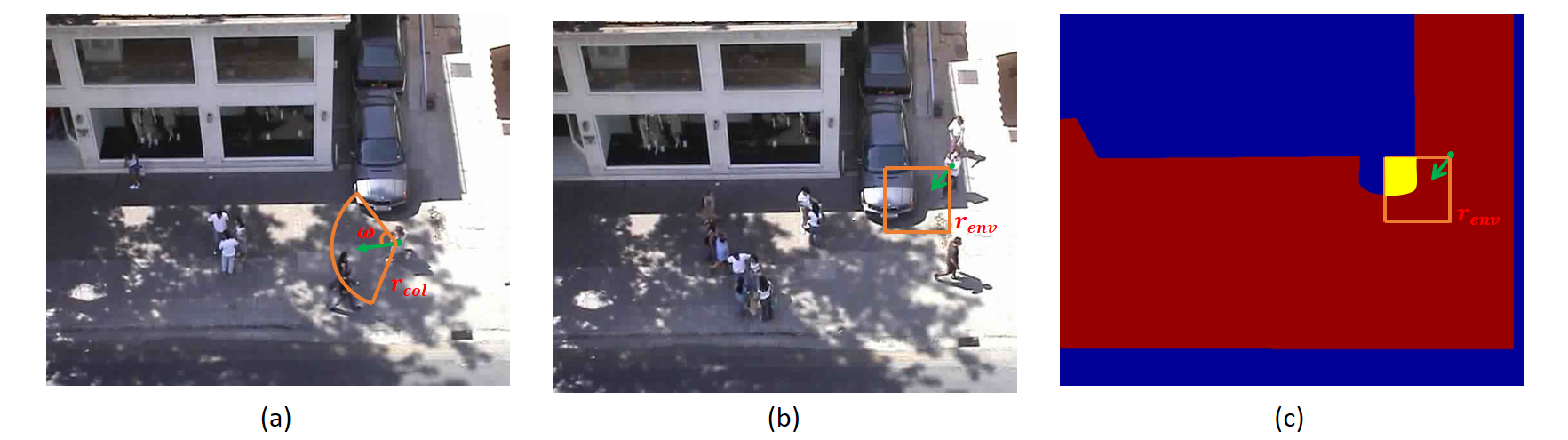}
\caption{(a) The neighborhood $\Omega(t)$ of a person is a sector within a circle (centered at this person with radius $r_{col}$) spanned by an angle $\omega$ from the current velocity vector (green arrow). (b) Each person has a view field (orange box) within which the environment repels a pedestrian. The view field is a square with dimension $r_{env}$ based on the current velocity vector (green arrow). The current velocity is along the diagonal of the orange box. (c) The environment is segmented into walkable (red) and unwalkable (blue) areas. Within the view field of the pedestrian in (b), the yellow pixels are the environment pixels that repel the pedestrian. $\omega$, $r_{col}$ and $r_{env}$ are hyperparameters.}
\label{fig:forceFields}
\end{figure}

\textbf{Inter-agent Repulsion}. Pedestrians often steer to avoid potential collisions and maintain personal space when other people are in the immediate neighborhood (\figref{forceFields} a). Given an agent $j$ in $\Omega_n^t$ of agent $n$ and her state $q_j^t$, agent $j$ repels agent $n$ based on $r_{nj} = p_n^t - p_j^t$:
\begin{equation}
\label{eq:collision}
    F_{col}^{nj} = -\nabla_{r_{nj}}\mathcal{U}_{nj}\left(\lVert r_{nj} \rVert\right), \text{ where }\mathcal{U}_{nj}\left(\lVert r_{nj} \rVert\right) = r_{col} k_{nj}e^{-\lVert r_{nj}\rVert/r_{col}} 
\end{equation}
where we employ a repulsive potential field $\mathcal{U}_{nj}\left(\lVert r_{nj}\rVert\right)$ modeled by a monotonic decreasing function of $\lVert r_{nj}\rVert$. Then the repulsive force caused by agent $j \in \Omega_n^t$ to agent $n$ is the gradient of $\mathcal{U}_{nj}$. Previously, simple functions such as symmetric elliptic fields were employed for $\mathcal{U}_{nj}$~\cite{helbing1995social}. Here, we model $\mathcal{U}_{nj}$ as a time-varying field parameterized by $k_{nj}$
which is learned via a neural network. Instead of directly learning $k_{nj}$, we set $k_{nj} = a*sigmoid(NN_{\phi_2}(q_n^t, q^t_{j, j\in \Omega_n^t})) + b$. $a$ and $b$ are hyperparameters to ensure that the learned $k_{nj}$ value is valid. If we have $m$ agents at time t in $\Omega_n^t$, the net repulsive force on agent $n$ is: $F_{col}^n = \sum_{j=0}^m F_{col}^{nj}$.

\textbf{Environment Repulsion}. Besides collisions with others, people also avoid nearby obstacles. We model the repulsion from the environment as:
\begin{equation}
\label{eq:env}
    F_{env} = \frac{k_{env}}{\lVert p_n^t - p_{obs}\rVert} (\frac{p_n^t - p_{obs}}{\lVert p_n^t - p_{obs}\rVert})
\end{equation}
where $p_{obs}$ is the position of the obstacle and $k_{env}$ is a learnable parameter. NSP-SFM learns $k_{env}$ directly via back-propagation and stochastic gradient descent. Since the environment is big, we assume the agent mainly focuses on her view field (\figref{forceFields} b) within which the environment (\figref{forceFields} c) repels the pedestrian. We calculate $p_{obs}$ as the center of the pixels that are classified as obstacles in the view field of an agent. $k_{env}$ is shared among all obstacles. So far, we have introduced all the interpretable parameters $\theta = \{\tau, k_{nj}, k_{env}\}$ in \Eqnref{nde}.

\textbf{Dynamics Stochasticity $\alpha(t, q^{t:t-M})$}. Trajectory prediction needs to explicitly model the motion randomness caused by intrinsic motion stochasticity and observational noises~\cite{wang_path_2016,wang_trending_2016}. 
We employ a more general setting by assuming the noise distribution can have arbitrary shapes and is also time varying, unlike previous formulations such as white noise~{\cite{He_Informative_2020}} which is too restrictive. Generally, learning such functions requires large amounts of data, as it is unconstrained. To constrain the learning, we further assume the noise is \textit{Normally} distributed in a latent space, rather than in the data space.

Given a prediction $\bar{p}^{t+1}$ without dynamics stochasticity and its corresponding observation $p^{t+1}$, there is an error $\alpha^{t+1} = \bar{p}^{t+1} - p^{t+1}$. To model the arbitrary and time-varying shape of the distribution of $\alpha^{t+1}$, we assume it depends on the brief history $p^{t:t-M}$ which implicitly considers the environment and other people. Then the conditional likelihood of $\alpha^{t+1}$ is: $P(\alpha^{t+1}|p^{t:t-M}) = \int P(\alpha^{t+1}|p^{t:t-M}, z)P(z)dz$, where $z$ is a latent variable. Assuming a mapping $Q(z | \alpha^{t+1}, p^{t:t-M})$ and $z$ being \textit{Normally} distributed, minimizing the KL divergence between $Q$, i.e., the variational posterior, and $P(z|\alpha^{t+1}, p^{t:t-M})$ leads to a conditional Variational Autoencoder (CVAE)~\cite{sohn_learning_2015}.


Our overall loss function is defined as $L = l_{traj} + l_{cvae}$ where:

\begin{align}
  l_{traj} =& \frac{1}{N(T-M)}\sum_{n=1}^N\sum_{t=M+1}^T\lVert p_n^t - \Bar{p}_n^t\rVert_2^2 \nonumber \\
  l_{cvae} =& \frac{1}{N(T-M)}\sum_{n=1}^N\sum_{t=M+1}^T\{ \lVert \alpha_n^t - \Tilde{\alpha}_n^t\rVert_2^2 \nonumber \\
  &+ \lambda D_{KL}(Q(z | \alpha_n^t, p^{t:t-M})||P(z| \alpha_n^t, p^{t:t-M}))\}
\end{align}
$N$ is the total number of samples, $M$ is the length of the history, and $T$ is the total length of the trajectory. $l_{traj}$ minimizes the difference between the predicted position and the ground-truth, while $l_{cvae}$ learns the distribution of randomness $\alpha$. During training, in each iteration, we assume the first $M+1$ frames of the trajectory are given and run the forward pass iteratively to predict the rest of the trajectory, then back-propagate to compute the gradient to update all parameters. During the forward pass, we use a semi-implicit scheme for stability: $\dot{p}^{t+1} = \dot{p}^t + \triangle t \ddot{p}^t$ and $p^{t+1} = p^t + \triangle t \dot{p}^{t+1}$.
We employ a progressive training scheme for the sub-nets. We first train Goal-Network with $l_{traj}$ only, then fix Goal-Network and add Collision-Network and $F_{env}$ for training using $l_{traj}$. Finally, we fix Goal-Network, Collision-Network and $F_{env}$, add $\alpha$ for training under $l_{cvae}$. We find this progressive training significantly improves the convergence speed. This is because we first train the deterministic part with the main forces added gradually, which converges quickly. Then the stochasticity part is trained separately to capture complex randomness. Please see the supplementary material for implementation details.

\subsection{NSP vs. Deep Neural Networks}
One big difference between NSP and existing deep learning is the deterministic system embedded in NSP. Instead of learning any function mapping the input to the output (as black box deep learning does), the deterministic system acts as a strong inductive bias and constrains the functional space within which the target mapping should lie. This is because a PDE family can be seen as a flow connecting the input and the output space~\cite{alvarez1999pde}, and the learning is essentially a process of finding the most fitting PDE within this flow. In addition to better data-fitting capability, this strong inductive bias also comes with two other advantages. First, the learned model can help explain motions because the PDE we employ is a physics system where the learnable parameters have physical meanings. Second, after learning, the PDE can be used to predict motions in drastically different scenes (e.g., with higher densities) and generate more plausible trajectories (e.g., fewer collisions). This is difficult for existing deep learning as it requires to extrapolate significantly to unseen interactions between pedestrians. 

\section{Experiments}
 
\subsection{Datasets}

We employ six widely used datasets in human trajectory prediction tasks: the Stanford Drone Dataset~\cite{robicquet2016learning}, ETH Hotel, ETH University~\cite{pellegrini2010improving}, UCY University, Zara1, and Zara2 datasets~\cite{lerner2007crowds}. \textbf{Stanford Drone Dataset (SDD):} SDD contains videos of a university campus with six classes of agents with rich interactions. SDD includes about 185,000 interactions between different agents and approximately 40,000 interactions between the agent and the environment. \textbf{ETH/UCY Datasets:} The datasets consist of human trajectories across five scenes recording the world coordinates of pedestrians. Following previous research~\cite{mangalam2021goals,mangalam2020not}, we adopt the standard leave-one-out evaluation protocol, where the model is trained on four sub-datasets and evaluated one. Since our goal sampling network and $F_{env}$ need to work in the pixel space, we project the world coordinates in ETH/UCY into the pixel space using the homography matrices provided in Y-net~\cite{mangalam2021goals}. When computing the prediction error, we project the predictions in the pixel space back into the world space. Finally, for SDD and ETH/UCY, we follow previous work~\cite{mangalam2021goals,sadeghian2018trajnet} to segment trajectories into 20-frame samples and split the dataset for training/testing. Given the first 8 ($M=7$) frames, we train NSP to predict the remaining 12 frames for each trajectory.

\subsection{Trajectory Prediction}
Average Displacement Error (ADE) and Final Displacement Error (FDE) are employed as previous research~\cite{alahi2016social,gupta2018social,mangalam2020not,mangalam2021goals}. ADE is calculated as the $l_2$ error between a predicted trajectory and the
ground truth, averaged over the entire trajectory. FDE is calculated as the $l_2$ error between the predicted final point and the ground truth. Following prior works, in the presence of multiple possible future predictions, the minimal error is reported. We compare our NSP-SFM with an extensive list of baselines, including published papers and unpublished technical reports: Social GAN (S-GAN)~\cite{gupta2018social}, Sophie~\cite{sadeghian2019sophie}, Conditional Flow VAE (CF-VAE)~\cite{bhattacharyya2019conditional}, Conditional Generative Neural System (CGNS)~\cite{li2019conditional}, NEXT~\cite{liang2019peeking}, P2TIRL~\cite{deo2020trajectory}, SimAug~\cite{liang2020simaug},  PECNet~\cite{mangalam2020not}, Traj++~\cite{salzmann2020trajectron++}, Multiverse~\cite{liang2020garden}, Y-Net~\cite{mangalam2021goals}, SIT~\cite{su2021pedestrian}, S-CSR~\cite{zhou2021sliding}, Social-DualCVAE~\cite{gao2022social} and CSCNet~\cite{xia2022cscnet}. We divide the baselines into two groups due to their setting differences. All baseline methods except S-CSR report the minimal error out of 20 sampled trajectories. S-CSR achieved better results by predicting 20 possible states in each step, and it is the only method adopting such sampling to our best knowledge. We refer to the former as standard-sampling and the latter as ultra-sampling. We compare NSP-SFM with S-CSR and other baseline methods under their respective settings.

Standard-sampling results are shown in \Tabref{sampling}. On SDD, NSP-SFM outperforms the best baseline Y-Net by 16.94\% and 10.46\% in ADE and FDE, respectively. In ETH/UCY, the improvement on average is 5.56\% and 11.11\% in ADE and FDE, with the maximal ADE improvement 12.5\% in UNIV and the maximal FDE improvement 27.27\% in ETH. We also compare NSP-SFM with S-CSR in \Tabref{ultraSampling}. NSP-SFM outperforms S-CSR on ETH/UCY by 70\% and 62.5\% on average in ADE and FDE. In SDD, the improvement is 35.74\% and 0.3\% (\Tabref{ultraSampling}). S-CSR is stochastic and learns per-step distributions, which enables it to draw 20 samples for every step during prediction. Therefore, the min error of S-CSR is much smaller than the other baselines. Similarly, NSP-SFM also learns a per-step distribution (the $\alpha$ function) despite its main behavior being dictated by a deterministic system. Under the same ultra-sampling setting, NSP-SFM outperforms S-CSR.

\begin{table}[tb]
\scriptsize
\begin{center}
\caption{Results on ETH/UCY and SDD based standard-sampling. NSP-SFM outperforms all baseline methods in both ADE and FDE. 20 samples are used in prediction and the minimal error is reported. $M=7$ in all experiments. The unit is meters on ETH/UCY and pixels on SDD.}
\begin{tabular}{ |p{2cm}|p{1cm}||p{0.9cm}|p{0.9cm}|p{0.9cm}|p{0.9cm}|p{0.9cm}|p{0.9cm}||p{0.9cm}| }
\hline
Methods& Metrics& ETH & Hotel &UNIV & ZARA1 & ZARA2 & AVG &SDD \\
\hline
\multirow{2}*{S-GAN~\cite{gupta2018social}} & ADE & 0.81 & 0.72 & 0.60 & 0.34 & 0.42 & 0.58 & 27.23 \\
& FDE & 1.52 & 1.61 & 1.26 & 0.69 & 0.84 & 1.18 & 41.44 \\
\hline
\multirow{2}*{Sophie~\cite{sadeghian2019sophie}} & ADE & 0.70 & 0.76 & 0.54 & 0.30 & 0.38 & 0.54 & 16.27 \\
& FDE & 1.43 & 1.67 & 1.24 & 0.63 & 0.78 & 1.15 & 29.38 \\
\hline
\multirow{2}*{CF-VAE~\cite{bhattacharyya2019conditional}} & ADE & N/A & N/A &  N/A & N/A &  N/A & N/A & 12.60 \\
& FDE & N/A & N/A &  N/A & N/A &  N/A & N/A & 22.30 \\
\hline
\multirow{2}*{CGNS~\cite{li2019conditional}} & ADE & 0.62 & 0.70 & 0.48 & 0.32 & 0.35 & 0.49 &15.6 \\
& FDE & 1.40 & 0.93 & 1.22 & 0.59 & 0.71 & 0.97 & 28.2 \\
\hline
\multirow{2}*{NEXT~\cite{liang2019peeking}} & ADE & 0.73 & 0.30 & 0.60 & 0.38 & 0.31 & 0.46 & N/A \\
& FDE & 1.65 & 0.59 & 1.27 & 0.81 & 0.68 & 1.00 & N/A \\
\hline
\multirow{2}*{P2TIRL~\cite{deo2020trajectory}} & ADE & N/A & N/A &  N/A & N/A &  N/A & N/A & 12.58 \\
& FDE & N/A & N/A &  N/A & N/A &  N/A & N/A & 22.07 \\
\hline
\multirow{2}*{SimAug~\cite{liang2020simaug}} & ADE & N/A & N/A &  N/A & N/A &  N/A & N/A & 10.27 \\
& FDE & N/A & N/A &  N/A & N/A &  N/A & N/A & 19.71 \\
\hline
\multirow{2}*{PECNet~\cite{mangalam2020not}} & ADE & 0.54 & 0.18 & 0.35 & 0.22 & 0.17 & 0.29 & 9.96 \\
& FDE & 0.87 & 0.24 & 0.60 & 0.39 & 0.30 & 0.48 & 15.88 \\
\hline
\multirow{2}*{Traj++~\cite{salzmann2020trajectron++}} & ADE & 0.39 & 0.12 & \textbf{0.20} & \textbf{0.15} & \textbf{0.11} & 0.19 & N/A \\
& FDE & 0.83 & 0.21 & 0.44 & 0.33 & 0.25 & 0.41 & N/A \\
\hline
\multirow{2}*{Multiverse~\cite{liang2020garden}} & ADE & N/A & N/A &  N/A & N/A &  N/A & N/A & 14.78 \\
 & FDE & N/A & N/A &  N/A & N/A &  N/A & N/A & 27.09 \\
\hline
\multirow{2}*{Y-net~\cite{mangalam2021goals}} & ADE & 0.28 & 0.10 & 0.24 & 0.17 & 0.13 & 0.18 &7.85 \\
& FDE & 0.33 & 0.14 & 0.41 & \textbf{0.27} & 0.22 & 0.27 & 11.85 \\
\hline
\multirow{2}*{SIT~\cite{su2021pedestrian}} & ADE & 0.38 & 0.11 & \textbf{0.20} & 0.16 & 0.12 & 0.19 & N/A \\
& FDE & 0.88 & 0.21 & 0.46 & 0.37 & 0.27 & 0.44 & N/A \\
\hline
Social & ADE & 0.66 & 0.34 & 0.39 & 0.27 & 0.24 & 0.38 & N/A \\
DualCVAE~\cite{gao2022social}& FDE & 1.18 & 0.61 & 0.74 & 0.48 & 0.42 & 0.69 & N/A \\
\hline
\multirow{2}*{CSCNet~\cite{xia2022cscnet}} & ADE & 0.51 & 0.22 & 0.36 & 0.31 & 0.47 & 0.37 & 14.63 \\
& FDE & 1.05 & 0.42 & 0.81 & 0.68 & 1.02 & 0.79 & 26.91 \\
\hline
NSP-SFM & ADE & \textbf{0.25} & \textbf{0.09} & 0.21 & 0.16 & 0.12 & \textbf{0.17} & \textbf{6.52} \\
(Ours)& FDE & \textbf{0.24} & \textbf{0.13} & \textbf{0.38} & \textbf{0.27} & \textbf{0.20} & \textbf{0.24} & \textbf{10.61} \\
\hline
\end{tabular}
\label{tab:sampling}
\end{center}
\end{table}

\begin{table}[tb]
\scriptsize
\begin{center}
\caption{Results on ETH/UCY (left) and SDD (right) based on ultra-sampling. 20 samples per step are used for prediction and the overall minimal error is reported. NSP-SFM outperforms S-CSR on both datasets in ADE and FDE.}
\begin{tabular}{ |p{1.6cm}|p{1cm}||p{0.9cm}|p{0.9cm}|p{0.9cm}|p{0.9cm}|p{0.9cm}|p{0.9cm}||p{0.9cm}| }
\hline
Methods & Metrics & ETH & Hotel & UNIV & ZARA1 & ZARA2 & Avg & SDD \\
\hline
\multirow{2}*{S-CSR~\cite{zhou2021sliding}} & ADE & 0.19 & 0.06 & 0.13 & 0.06 & 0.06 & 0.10 & 2.77 \\
& FDE & 0.35 & \textbf{0.07} & 0.21 & 0.07 & 0.08 & 0.16 & 3.45 \\
\hline
\multirow{2}*{NSP-SFM} & ADE & \textbf{0.07} & \textbf{0.03} & \textbf{0.03} & \textbf{0.02} & \textbf{0.02} & \textbf{0.03} & \textbf{1.78} \\
& FDE & \textbf{0.09} & \textbf{0.07} & \textbf{0.04} & \textbf{0.04} & \textbf{0.04} & \textbf{0.06} & \textbf{3.44} \\
\hline
\end{tabular}
\label{tab:ultraSampling}
\end{center}
\end{table}

\subsection{Generalization to Unseen Scenarios}
We evaluate NSP-SFM on significantly different scenarios after training. We increase the scene density as it is a major factor in pedestrian dynamics~\cite{narang2015generating}. This is through randomly sampling initial and goal positions and let NSP-SFM predict the trajectories. Since there is no ground truth, to evaluate the prediction plausibility, we employ collision rate because it is widely adopted~\cite{liu2021social} and parsimonious: regardless of the specific behaviors of agents, they do not penetrate each other in the real world. The collision rate is computed based on the percentage of trajectories colliding with one another. We treat each agent as a disc with radius $r=0.2 \; m$ in ECY/UCY and $r=15$ pixels in SDD. Once the distance between two agents falls below $2r$, we count the two trajectories as in collision. Due to the tracking error and the distorted images, the ground truth $r$ is hard to obtain. We need to estimate $r$. If it is too large, the collision rate will be high in all cases; otherwise the collision rate will be too low, e.g., $r=0$ will give $0\%$ collision rate all the time. Therefore, we did a search and found that the above values are reasonable as they keep the collision rate of the ground-truth data approximately zero. We show two experiments. The first is the collision rate on the testing data, and the second is scenarios with higher densities. While the first is mainly to compare the plausibility of the prediction, the second is to test the model generalizability. For comparison, we choose two state-of-the-art baseline methods: Y-net and S-CSR. Y-net is published which achieves the best performance, while S-CSR is unpublished but claims to achieve better performance.

\begin{table}[b]
\begin{center}
\caption{Collision rate on testing data in ETH/UCY and SDD. NSP-SFM universally outperforms all baseline methods.}
\begin{tabular}{ |p{1.5cm}||p{1.1cm}|p{1.1cm}|p{1.5cm}|p{1.5cm}|p{1.5cm}|p{1.3cm}| |p{1.3cm}|}
 \hline
  Methods& ETH & Hotel & UNIV & ZARA1 & ZARA2 & Avg & SDD \\
 \hline
 Y-net & 0 & 0 & 1.51\% & 0.82\% & 1.31\% & 0.73\% & 0.47\% \\
 \hline
 S-CSR & 0 & 0  & 1.82\% & 0.41\% & 1.31\% & 0.71\% & \textbf{0.42\%} \\
 \hline
 NSP-SFM & \textbf{0} & \textbf{0} & \textbf{1.48\%} & \textbf{0} & \textbf{0.66\%} & \textbf{0.43\%} & \textbf{0.42\%}\\
 \hline
\end{tabular}
\label{tab:collisionETH}
\end{center}
\end{table}

\begin{table}[tb]
\scriptsize 
\begin{center}
\caption{Collision rates of the generalization experiments on ZARA2 (Z) and coupa0 (C). NSP-SFM shows strong generalizability in unseen high density scenarios.}
\begin{tabular}{ |p{1.3cm}||p{1cm}|p{1cm}|p{1cm}|p{1.2cm}||p{1cm}|p{1cm}|p{1cm}| p{1.2cm}| }
 \hline
 Methods& Z(1) & Z(2) & Z(3) & Z(avg)& C(1) & C(2) & C(3)& C(avg)\\
 \hline
 Y-net  & 1.8\% & 2.2\% & 2.0\% &  2.0\% & 2.8\% & 2.9\% & 3.8\%& 3.2\% \\
 S-CSR  & 3.2\% & 2.4\% & 1.8\% & 2.5\%  & 2.5\% & 1.7\% & 1.9\% & 2.0\% \\
 NSP-SFM  & \textbf{0.2\%} & \textbf{0.2\%} & \textbf{0} & \textbf{0.1\%}  & \textbf{0.6\%} & \textbf{0.6\%} & \textbf{0.6\%} & \textbf{0.6\%} \\
 \hline
\end{tabular}
\label{tab:collisionSim}
\end{center}
\end{table}

\Tabref{collisionETH} shows the comparison of the collision rate. NSP-SFM outperforms the baseline methods in generating trajectories with fewer collisions. Y-net and S-CSR also perform well on the testing data because their predictions are close to the ground-truth. Nevertheless, they are still worse than NSP-SFM. Next, we test drastically different scenarios. We use ZARA2 and coupa0 (a sub-dataset from SDD) as the environment and randomly sample the initial positions and goals for 32 and 50 agents respectively. Because the highest number of people that simultaneously appear in the scene is 14 in ZARA2 and 11 in coupa0, we effectively increase the density by 2-5 times. For NSP-SFM, the initial and goal positions are sufficient. For Y-net and S-CSR which require 8 frames (3.2 Seconds) as input, we use NSP-SFM to simulate the first 8 frames of each agent, then feed them into both baselines. \Tabref{collisionSim} shows the results of three experiments. Since the density is significantly higher than the data, both Y-net and S-CSR cause much higher collision rate. While NSP-SFM's collision rate also occasionally increases (i.e. SDD) compared with \Tabref{collisionETH}, it is far more plausible.

\subsection{Interpretability of Prediction}
\begin{figure}[tb]
\centering
\includegraphics[width=0.8\textwidth]{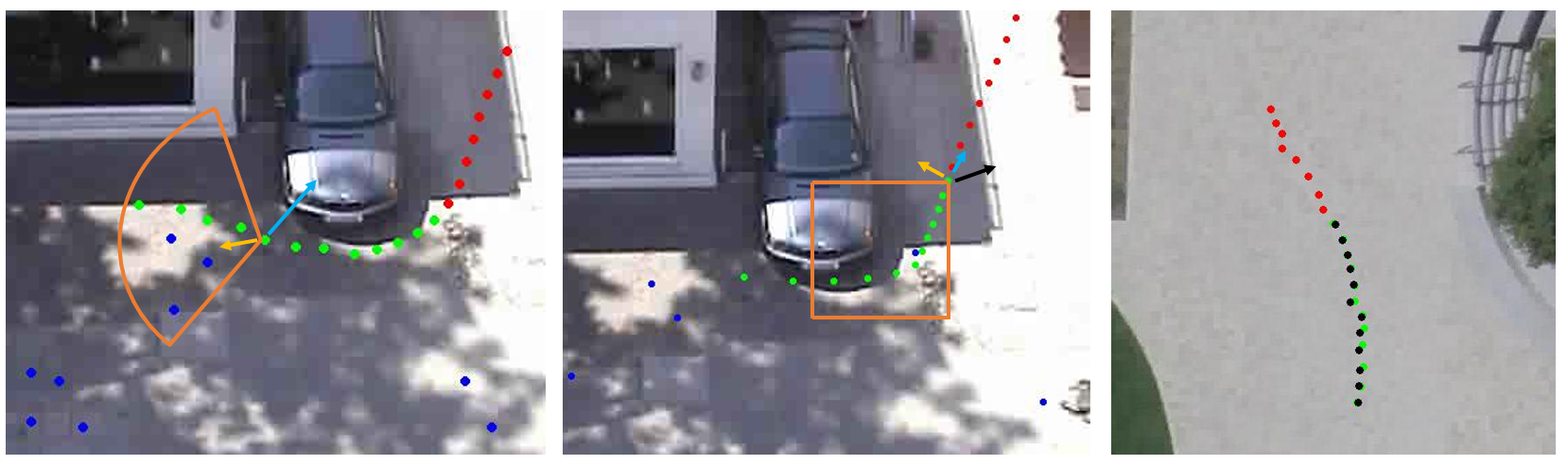}
\caption{Red dots are observed, green dots are our prediction and black dots are the ground-truth. Blue dots are pedestrians. $F_{goal}$, $F_{col}$ and $F_{env}$ are shown as yellow, light blue and black arrows for a person. The orange areas are the view field for avoiding collisions with other people (left) and the environment (middle). They provide plausible explanations of individual behaviors such as steering. Left and middle show the major influence of different forces. Right shows motion randomness captured by our model.}
\label{fig:interpretability}
\end{figure}

Unlike black-box deep learning methods, NSP-SFM has an embedded explainable model. While predicting a trajectory, NSP can also provide plausible explanations of the motion, by estimating the `forces' exerted on a specific person. This potentially enables NSP-SFM to be used in applications beyond prediction, e.g. behavior analysis~\cite{zeng2014application}. \Figref{interpretability} Left shows that a person, instead of directly walking towards the goal, steered upwards (the green trajectory in the orange area). This could be explained by the strong repulsive force (the light blue arrow) which is generated by the potential collisions with the agents in front of this person, in line with existing studies~\cite{narang2015generating}. Similar explanations can be made in \Figref{interpretability} Middle, where all three forces are present. $F_{env}$ (the black arrow) is the most prominent, as expected, as the person is very close to the car. The repulsive force (light blue arrow) also plays a role due to the person in front of the agent (the blue dot in the orange area). 

\Figref{interpretability} Right shows an example where motion randomness is captured by NSP. In this example, there was no other pedestrian and the person was not close to any obstacle. However, the trajectory still significantly deviates from a straight line, which cannot be fully explained by e.g. the principle of minimal energy expenditure~\cite{virtanen2013energy}. The deviation could be caused by unobserved factors, e.g. the agent changing her goal or being distracted by something on the side. These factors do not only affect the trajectory but also the dynamics, e.g. sudden changes of velocity. These unobserved random factors are implicitly captured by the CVAE in NSP-SFM. More results are in the supplementary material.

We emphasize that NSP-SFM merely provides plausible explanations and by no means the only possible explanations. Although explaining behaviors based on physics models has been widely used, there can be alternative explanations~\cite{wang2016understanding}. Visualizing the forces is merely one possible way. Theoretically, it is also possible to visualize deep neural networks, e.g. layer activation. However, it is unclear how or which layer to visualize to explain the motion.  Overall, NSP-SFM is more explainable than black-box deep learning. 

\subsection{Ablation Study}

\begin{table}[tb]
\begin{center}
\caption{Ablation study on SDD. (w/o) means without CVAE and (w) means with CVAE. $F_{goal}$ is goal attraction only and NSP-SFM is all three forces.}
\begin{tabular}{ |p{1.2cm}||p{2cm}|p{2.5cm}|p{2cm}| }
 \hline
 SDD & $F_{goal}$(w/o) & NSP-SFM(w/o) & NSP-SFM(w)\\
 \hline
 ADE & 6.57 & 6.52 & 1.78\\
 FDE & 10.68 & 10.61 & 3.44\\
 \hline
\end{tabular}
\label{tab:ablation}
\end{center}
\end{table}

\begin{figure}[tb]
\centering
\includegraphics[width=\textwidth]{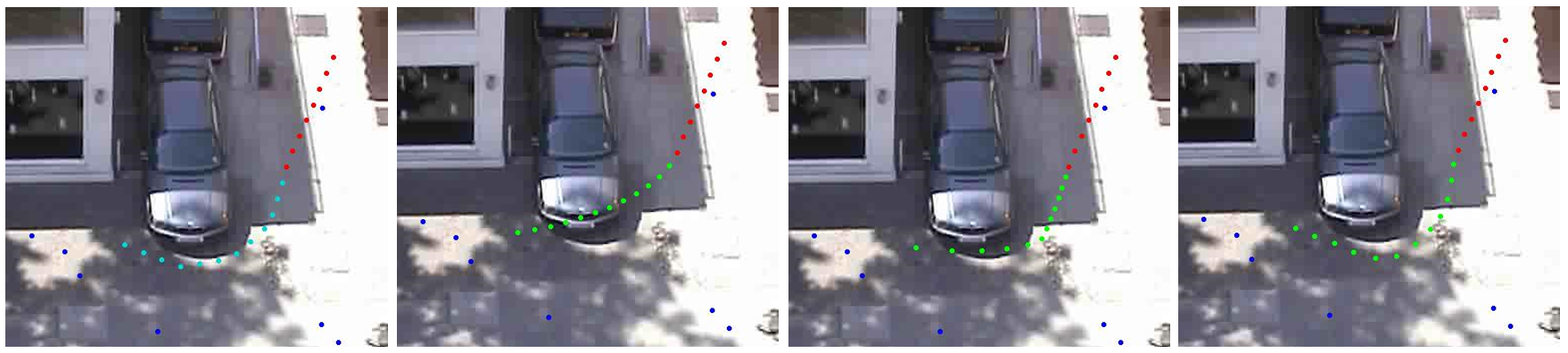}
\caption{Red, green and cyan dots are observations, prediction and ground-truth respectively. From left to right: ground truth, $F_{goal}$(w/o), NSP-SFM(w/o) and NSP-SFM(w).}
\label{fig:ablation}
\end{figure}

To further investigate the roles of different components, we conduct an ablation study on SDD with three settings: $F_{goal}$(w/o) with goal attraction only, i.e. omitting other components such as $F_{col}$, $F_{env}$ and dynamics stochasticity; NSP-SFM (w/o) without dynamics stochasticity; and NSP-SFM (w) the full model. The results are shown in \Tabref{ablation}. Interestingly, $F_{goal}$(w/o) can already achieve good results. This is understandable as it is trained first in our progressive training scheme and catches most of the dynamics. NSP-SFM (w/o) further improves the performance. The improvement seems to be small but we find the other repulsive forces are crucial for trajectories with irregular geometries such as avoiding obstacles. Further NSP-SFM (w) significantly improves the results because it enables NSP to learn the dynamics stochasticity via a per-step distribution. We show one example in \Figref{ablation} in all settings. More ablation experiments can be found in the supplementary material.

\section{Conclusions, Limitations, and Future Work}

In this paper, we have proposed a new Neural Differential Equation model for trajectory prediction. Through exhaustive evaluation and comparison, our model, Neural Social Physics, has proven to be more accurate in trajectory prediction, generalize well in significantly different scenarios and can provide possible explanations for motions. The major limitation of NSP lies in the physics model, which overly simplifies people into 2D particles. In real-world scenarios, people are much more complex, and their motions can be influenced by other factors such as their affective states or interact with dense scenarios~\cite{bera2014realtime,bera2017aggressive}. It would be useful to extend our NSP framework by incorporating these ideas and handle complex  systems such as fluids/fields/agent-based modeling can be adopted to replace the components in \Eqnref{timeDiscretization}. In the future, we would like to extend the current framework to model high-density crowds, where continuum models or reciprocal  velocity obstacles need to be used. We would also like to incorporate learning-based collision detection techniques into this framework~\cite{tan2021lcollision,tan2022n}.

\paragraph{Acknowledgements} This project has received funding from the European Union’s Horizon 2020 research and innovation programme under grant agreement No 899739 CrowdDNA.

\bibliographystyle{splncs04}
\bibliography{egbib}

\begin{thebibliography}{10}
\providecommand{\url}[1]{\texttt{#1}}
\providecommand{\urlprefix}{URL }
\providecommand{\doi}[1]{https://doi.org/#1}

\bibitem{alahi2016social}
Alahi, A., Goel, K., Ramanathan, V., Robicquet, A., Fei-Fei, L., Savarese, S.:
  Social lstm: Human trajectory prediction in crowded spaces. In: Proceedings
  of the IEEE conference on computer vision and pattern recognition. pp.
  961--971 (2016)

\bibitem{alvarez1999pde}
{\'A}lvarez~Le{\'o}n, L.M., Esclar{\'\i}n~Monreal, J., Lef{\'e}bure, M.,
  S{\'a}nchez, J.: A pde model for computing the optical flow  (1999)

\bibitem{antonucci2020generating}
Antonucci, A., Papini, G.P.R., Palopoli, L., Fontanelli, D.: Generating
  reliable and efficient predictions of human motion: A promising encounter
  between physics and neural networks. arXiv preprint arXiv:2006.08429  (2020)

\bibitem{bartoli2018context}
Bartoli, F., Lisanti, G., Ballan, L., Del~Bimbo, A.: Context-aware trajectory
  prediction. In: 2018 24th International Conference on Pattern Recognition
  (ICPR). pp. 1941--1946. IEEE (2018)

\bibitem{BENDALIBRAHAM_recent_2021}
Bendali-Braham, M., Weber, J., Forestier, G., Idoumghar, L., Muller, P.A.:
  Recent trends in crowd analysis: A review. Machine Learning with Applications
   \textbf{4},  100023 (2021)

\bibitem{bera2014realtime}
Bera, A., Manocha, D.: Realtime multilevel crowd tracking using reciprocal
  velocity obstacles. In: 2014 22nd International Conference on Pattern
  Recognition. pp. 4164--4169. IEEE (2014)

\bibitem{bera2017aggressive}
Bera, A., Randhavane, T., Manocha, D.: Aggressive, tense or shy? identifying
  personality traits from crowd videos. In: IJCAI. pp. 112--118 (2017)

\bibitem{vanDenBerg_reciprocal_2008}
van~den Berg, J., Lin, M., Manocha, D.: Reciprocal velocity obstacles for
  real-time multi-agent navigation. In: 2008 IEEE International Conference on
  Robotics and Automation (2008)

\bibitem{bhattacharyya2019conditional}
Bhattacharyya, A., Hanselmann, M., Fritz, M., Schiele, B., Straehle, C.N.:
  Conditional flow variational autoencoders for structured sequence prediction.
  In: 4th Workshop on Bayesian Deep Learning. bayesiandeeplearning. org (2019)

\bibitem{cai2022physics}
Cai, S., Mao, Z., Wang, Z., Yin, M., Karniadakis, G.E.: Physics-informed neural
  networks (pinns) for fluid mechanics: A review. Acta Mechanica Sinica pp.
  1--12 (2022)

\bibitem{chaker2017social}
Chaker, R., Al~Aghbari, Z., Junejo, I.N.: Social network model for crowd
  anomaly detection and localization. Pattern Recognition  \textbf{61},
  266--281 (2017)

\bibitem{charalambous2014data}
Charalambous, P., Karamouzas, I., Guy, S.J., Chrysanthou, Y.: A data-driven
  framework for visual crowd analysis. In: Computer Graphics Forum. vol.~33,
  pp. 41--50. Wiley Online Library (2014)

\bibitem{chen2018neural}
Chen, R.T., Rubanova, Y., Bettencourt, J., Duvenaud, D.K.: Neural ordinary
  differential equations. Advances in neural information processing systems
  \textbf{31} (2018)

\bibitem{deo2020trajectory}
Deo, N., Trivedi, M.M.: Trajectory forecasts in unknown environments
  conditioned on grid-based plans. arXiv preprint arXiv:2001.00735  (2020)

\bibitem{ellis2009modelling}
Ellis, D., Sommerlade, E., Reid, I.: Modelling pedestrian trajectory patterns
  with gaussian processes. In: 2009 IEEE 12th International Conference on
  Computer Vision Workshops, ICCV Workshops. pp. 1229--1234. IEEE (2009)

\bibitem{gao2022social}
Gao, J., Shi, X., Yu, J.J.: Social-dualcvae: Multimodal trajectory forecasting
  based on social interactions pattern aware and dual conditional variational
  auto-encoder. arXiv preprint arXiv:2202.03954  (2022)

\bibitem{Gong_finegrained_2022}
Gong, D., Zhu, Z., Andrew, B., Wang, H.: Fine-grained differentiable physics: a
  yarn-level model for fabrics. In: International Conference on Learning
  Representations (2022)

\bibitem{gupta2018social}
Gupta, A., Johnson, J., Fei-Fei, L., Savarese, S., Alahi, A.: Social gan:
  Socially acceptable trajectories with generative adversarial networks. In:
  Proceedings of the IEEE conference on computer vision and pattern
  recognition. pp. 2255--2264 (2018)

\bibitem{He_Informative_2020}
He, F., Xia, Y., Zhao, X., Wang, H.: Informative scene decomposition for crowd
  analysis, comparison and simulation guidance. ACM Transaction on Graphics
  (TOG)  \textbf{4}(39) (2020)

\bibitem{helbing1995social}
Helbing, D., Molnar, P.: Social force model for pedestrian dynamics. Physical
  review E  \textbf{51}(5), ~4282 (1995)

\bibitem{hossain2022sfmgnet}
Hossain, S., Johora, F.T., M{\"u}ller, J.P., Hartmann, S., Reinhardt, A.:
  Sfmgnet: A physics-based neural network to predict pedestrian trajectories.
  arXiv  (2022)

\bibitem{ivanovic2019trajectron}
Ivanovic, B., Pavone, M.: The trajectron: Probabilistic multi-agent trajectory
  modeling with dynamic spatiotemporal graphs. In: Proceedings of the IEEE/CVF
  International Conference on Computer Vision. pp. 2375--2384 (2019)

\bibitem{karamouzas2018crowd}
Karamouzas, I., Sohre, N., Hu, R., Guy, S.J.: Crowd space: a predictive crowd
  analysis technique. ACM Transactions on Graphics (TOG)  \textbf{37}(6),
  1--14 (2018)

\bibitem{Karniadakis_datadriven_2021}
Karniadakis, G.E., Kevrekidis, I.G., Lu, L., Perdikaris, P., Wang, S., Yang,
  L.: Physics-informed machine learning. Nat Rev Phys (3),  422–440 (2021)

\bibitem{kidger_neural_2022}
Kidger, P.: On neural differential equations (2022)

\bibitem{kim2015interactive}
Kim, S., Bera, A., Manocha, D.: Interactive crowd content generation and
  analysis using trajectory-level behavior learning. In: 2015 IEEE
  International Symposium on Multimedia (ISM). pp. 21--26. IEEE (2015)

\bibitem{kreiss2021deep}
Kreiss, S.: Deep social force. arXiv preprint arXiv:2109.12081  (2021)

\bibitem{lerner2007crowds}
Lerner, A., Chrysanthou, Y., Lischinski, D.: Crowds by example. In: Computer
  graphics forum. vol.~26, pp. 655--664. Wiley Online Library (2007)

\bibitem{li2019conditional}
Li, J., Ma, H., Tomizuka, M.: Conditional generative neural system for
  probabilistic trajectory prediction. In: 2019 IEEE/RSJ International
  Conference on Intelligent Robots and Systems (IROS). pp. 6150--6156. IEEE
  (2019)

\bibitem{liang2019differentiable}
Liang, J., Lin, M., Koltun, V.: Differentiable cloth simulation for inverse
  problems. Advances in Neural Information Processing Systems  \textbf{32}
  (2019)

\bibitem{liang2020simaug}
Liang, J., Jiang, L., Hauptmann, A.: Simaug: Learning robust representations
  from 3d simulation for pedestrian trajectory prediction in unseen cameras.
  arXiv preprint arXiv:2004.02022  \textbf{2} (2020)

\bibitem{liang2020garden}
Liang, J., Jiang, L., Murphy, K., Yu, T., Hauptmann, A.: The garden of forking
  paths: Towards multi-future trajectory prediction. In: Proceedings of the
  IEEE/CVF Conference on Computer Vision and Pattern Recognition. pp.
  10508--10518 (2020)

\bibitem{liang2019peeking}
Liang, J., Jiang, L., Niebles, J.C., Hauptmann, A.G., Fei-Fei, L.: Peeking into
  the future: Predicting future person activities and locations in videos. In:
  Proceedings of the IEEE/CVF Conference on Computer Vision and Pattern
  Recognition. pp. 5725--5734 (2019)

\bibitem{liu2021social}
Liu, Y., Yan, Q., Alahi, A.: Social nce: Contrastive learning of socially-aware
  motion representations. In: Proceedings of the IEEE/CVF International
  Conference on Computer Vision. pp. 15118--15129 (2021)

\bibitem{lopez2019character}
L{\'o}pez, A., Chaumette, F., Marchand, E., Pettr{\'e}, J.: Character
  navigation in dynamic environments based on optical flow. In: Computer
  Graphics Forum. vol.~38, pp. 181--192. Wiley Online Library (2019)

\bibitem{Luo_Agentbased_2008}
Luo, L., Zhou, S., Cai, W., Low, M.Y.H., Tian, F., Wang, Y., Xiao, X., Chen,
  D.: Agent‐based human behavior modeling for crowd simulation. Computer
  Animation and Virtual Worlds  \textbf{19} (2008)

\bibitem{mangalam2021goals}
Mangalam, K., An, Y., Girase, H., Malik, J.: From goals, waypoints \& paths to
  long term human trajectory forecasting. In: Proceedings of the IEEE/CVF
  International Conference on Computer Vision. pp. 15233--15242 (2021)

\bibitem{mangalam2020not}
Mangalam, K., Girase, H., Agarwal, S., Lee, K.H., Adeli, E., Malik, J., Gaidon,
  A.: It is not the journey but the destination: Endpoint conditioned
  trajectory prediction. In: European Conference on Computer Vision. pp.
  759--776. Springer (2020)

\bibitem{mohamed2020social}
Mohamed, A., Qian, K., Elhoseiny, M., Claudel, C.: Social-stgcnn: A social
  spatio-temporal graph convolutional neural network for human trajectory
  prediction. In: Proceedings of the IEEE/CVF Conference on Computer Vision and
  Pattern Recognition. pp. 14424--14432 (2020)

\bibitem{narain2009aggregate}
Narain, R., Golas, A., Curtis, S., Lin, M.C.: Aggregate dynamics for dense
  crowd simulation. In: ACM SIGGRAPH Asia 2009 papers, pp.~1--8 (2009)

\bibitem{narang2015generating}
Narang, S., Best, A., Curtis, S., Manocha, D.: Generating pedestrian
  trajectories consistent with the fundamental diagram based on physiological
  and psychological factors. PLoS one  \textbf{10}(4),  e0117856 (2015)

\bibitem{oliver2000bayesian}
Oliver, N.M., Rosario, B., Pentland, A.P.: A bayesian computer vision system
  for modeling human interactions. IEEE transactions on pattern analysis and
  machine intelligence  \textbf{22}(8),  831--843 (2000)

\bibitem{pellegrini2010improving}
Pellegrini, S., Ess, A., Gool, L.V.: Improving data association by joint
  modeling of pedestrian trajectories and groupings. In: European conference on
  computer vision. pp. 452--465. Springer (2010)

\bibitem{rackauckas2020universal}
Rackauckas, C., Ma, Y., Martensen, J., Warner, C., Zubov, K., Supekar, R.,
  Skinner, D., Ramadhan, A., Edelman, A.: Universal differential equations for
  scientific machine learning. arXiv preprint arXiv:2001.04385  (2020)

\bibitem{raissi2019physics}
Raissi, M., Perdikaris, P., Karniadakis, G.E.: Physics-informed neural
  networks: A deep learning framework for solving forward and inverse problems
  involving nonlinear partial differential equations. Journal of Computational
  physics  \textbf{378},  686--707 (2019)

\bibitem{robicquet2016learning}
Robicquet, A., Sadeghian, A., Alahi, A., Savarese, S.: Learning social
  etiquette: Human trajectory understanding in crowded scenes. In: European
  conference on computer vision. pp. 549--565. Springer (2016)

\bibitem{ronneberger2015u}
Ronneberger, O., Fischer, P., Brox, T.: U-net: Convolutional networks for
  biomedical image segmentation. In: International Conference on Medical image
  computing and computer-assisted intervention. pp. 234--241. Springer (2015)

\bibitem{sadeghian2018trajnet}
Sadeghian, A., Kosaraju, V., Gupta, A., Savarese, S., Alahi, A.: Trajnet:
  Towards a benchmark for human trajectory prediction. arXiv preprint  (2018)

\bibitem{sadeghian2019sophie}
Sadeghian, A., Kosaraju, V., Sadeghian, A., Hirose, N., Rezatofighi, H.,
  Savarese, S.: Sophie: An attentive gan for predicting paths compliant to
  social and physical constraints. In: Proceedings of the IEEE/CVF Conference
  on Computer Vision and Pattern Recognition. pp. 1349--1358 (2019)

\bibitem{salzmann2020trajectron++}
Salzmann, T., Ivanovic, B., Chakravarty, P., Pavone, M.: Trajectron++:
  Dynamically-feasible trajectory forecasting with heterogeneous data. In:
  European Conference on Computer Vision. pp. 683--700. Springer (2020)

\bibitem{Shen_high_2021}
Shen, S., Yang, Y., Shao, T., Wang, H., Jiang, C., Lan, L., Zhou, K.:
  High-order differentiable autoencoder for nonlinear model reduction. ACM
  Trans. Graph.  \textbf{40}(4) (jul 2021)

\bibitem{shen2018data}
Shen, Y., Henry, J., Wang, H., Ho, E.S.L., Komura, T., Shum, H.P.H.:
  {Data‐Driven Crowd Motion Control With Multi‐Touch Gestures}. Computer
  Graphics Forum  (2018). \doi{10.1111/cgf.13333}

\bibitem{shi2021sgcn}
Shi, L., Wang, L., Long, C., Zhou, S., Zhou, M., Niu, Z., Hua, G.: Sgcn: Sparse
  graph convolution network for pedestrian trajectory prediction. In:
  Proceedings of the IEEE/CVF Conference on Computer Vision and Pattern
  Recognition. pp. 8994--9003 (2021)

\bibitem{sighencea2021review}
Sighencea, B.I., Stanciu, R.I., C{\u{a}}leanu, C.D.: A review of deep
  learning-based methods for pedestrian trajectory prediction. Sensors
  \textbf{21}(22), ~7543 (2021)

\bibitem{sohn_learning_2015}
Sohn, K., Lee, H., Yan, X.: Learning structured output representation using
  deep conditional generative models. Advances in neural information processing
  systems  \textbf{28} (2015)

\bibitem{su2021pedestrian}
Su, T., Meng, Y., Xu, Y.: Pedestrian trajectory prediction via spatial
  interaction transformer network. In: 2021 IEEE Intelligent Vehicles Symposium
  Workshops (IV Workshops). pp. 154--159. IEEE (2021)

\bibitem{tan2021lcollision}
Tan, Q., Pan, Z., Manocha, D.: Lcollision: Fast generation of collision-free
  human poses using learned non-penetration constraints. In: Proceedings of the
  AAAI Conference on Artificial Intelligence. vol.~35, pp. 3913--3921 (2021)

\bibitem{tan2022n}
Tan, Q., Pan, Z., Smith, B., Shiratori, T., Manocha, D.: N-penetrate: Active
  learning of neural collision handler for complex 3d mesh deformations. In:
  International Conference on Machine Learning. pp. 21037--21049. PMLR (2022)

\bibitem{vantoll_algorithms_2021}
Van~Toll, W., Pettr{\'e}, J.: {Algorithms for Microscopic Crowd Simulation:
  Advancements in the 2010s}. {Computer Graphics Forum}  \textbf{40}(2) (2021)

\bibitem{vemula2018social}
Vemula, A., Muelling, K., Oh, J.: Social attention: Modeling attention in human
  crowds. In: 2018 IEEE international Conference on Robotics and Automation
  (ICRA). pp. 4601--4607. IEEE (2018)

\bibitem{virtanen2013energy}
Virtanen, A.: Energy-based pedestrian navigation. In: Proc. 20th ITS World
  Congr. pp.~1--9 (2013)

\bibitem{wan2017learning}
Wan, Z., Hu, X., He, H., Guo, Y.: A learning based approach for social force
  model parameter estimation. In: IJCNN. pp. 4058--4064. IEEE (2017)

\bibitem{wang_path_2016}
Wang, H., Ond{\v{r}}ej, J., O'Sullivan, C.: Path patterns: Analyzing and
  comparing real and simulated crowds. In: ACM SIGGRAPH Symposium on
  Interactive 3D Graphics and Games 2016. pp. 49--57 (2016)

\bibitem{wang_trending_2016}
Wang, H., Ond{\v{r}}ej, J., O'Sullivan, C.: Trending paths: A new
  semantic-level metric for comparing simulated and real crowd data. IEEE
  Transactions on Visualization and Computer Graphics  \textbf{PP}(99), ~1--1
  (2016)

\bibitem{wang2016globally}
Wang, H., O’Sullivan, C.: Globally continuous and non-markovian crowd
  activity analysis from videos. In: European conference on computer vision.
  pp. 527--544. Springer (2016)

\bibitem{wang2016understanding}
Wang, P.: Understanding social-force model in psychological principles of
  collective behavior. arXiv preprint arXiv:1605.05146  (2016)

\bibitem{wang2011trajectory}
Wang, X., Ma, K.T., Ng, G.W., Grimson, W.E.L.: Trajectory analysis and semantic
  region modeling using nonparametric hierarchical bayesian models.
  International journal of computer vision  \textbf{95}(3),  287--312 (2011)

\bibitem{wei2020simulating}
Wei, J., Fan, W., Li, Z., Guo, Y., Fang, Y., Wang, J.: Simulating crowd
  evacuation in a social force model with iterative extended state observer.
  Journal of advanced transportation  \textbf{2020} (2020)

\bibitem{Werling2021fast}
Werling, K., Omens, D., Lee, J., Exarchos, I., Liu, C.K.: Fast and
  feature-complete differentiable physics for articulated rigid bodies with
  contact. CoRR  \textbf{abs/2103.16021} (2021)

\bibitem{Wolinski_paramter_2014}
Wolinski, D., J.~Guy, S., Olivier, A.H., Lin, M., Manocha, D., Pettré, J.:
  Parameter estimation and comparative evaluation of crowd simulations.
  Computer Graphics Forum  \textbf{33}(2),  303--312 (2014)

\bibitem{xia2022cscnet}
Xia, B., Wong, C., Peng, Q., Yuan, W., You, X.: Cscnet: Contextual semantic
  consistency network for trajectory prediction in crowded spaces. Pattern
  Recognition p. 108552 (2022)

\bibitem{zeng2014application}
Zeng, W., Chen, P., Nakamura, H., Iryo-Asano, M.: Application of social force
  model to pedestrian behavior analysis at signalized crosswalk. Transportation
  research part C: emerging technologies  \textbf{40},  143--159 (2014)

\bibitem{zhang_meshingnet3d_2021}
Zhang, Z., Jimack, P.K., Wang, H.: {MeshingNet3D}: {Efficient} generation of
  adapted tetrahedral meshes for computational mechanics. Advances in
  Engineering Software  \textbf{157-158} (Jul 2021)

\bibitem{Zhang_MeshingNet_2020}
Zhang, Z., Wang, Y., Jimack, P.K., Wang, H.: Meshingnet: A new mesh generation
  method based on deep learning. In: Computational Science -- ICCS 2020. pp.
  186--198. Springer International Publishing, Cham (2020)

\bibitem{zhong2019symplectic}
Zhong, Y.D., Dey, B., Chakraborty, A.: Symplectic ode-net: Learning hamiltonian
  dynamics with control. arXiv preprint arXiv:1909.12077  (2019)

\bibitem{zhou2011random}
Zhou, B., Wang, X., Tang, X.: Random field topic model for semantic region
  analysis in crowded scenes from tracklets. In: CVPR 2011. pp. 3441--3448.
  IEEE (2011)

\bibitem{zhou2021sliding}
Zhou, H., Ren, D., Yang, X., Fan, M., Huang, H.: Sliding sequential cvae with
  time variant socially-aware rethinking for trajectory prediction. arXiv
  preprint arXiv:2110.15016  (2021)

\bibitem{zubov2021NeuralPDE}
Zubov, K., McCarthy, Z., Ma, Y., Calisto, F., Pagliarino, V., Azeglio, S.,
  Bottero, L., Luj{\'{a}}n, E., Sulzer, V., Bharambe, A., Vinchhi, N.,
  Balakrishnan, K., Upadhyay, D., Rackauckas, C.: Neuralpde: Automating
  physics-informed neural networks (pinns) with error approximations. CoRR
  \textbf{abs/2107.09443} (2021)

\end{thebibliography}

\newpage
\begin{table}[!b]
\centering
\caption{Collision rates of the generalization experiments on Coupa0. Results of Y-net, S-SCR and NSP on 50, 74, 100, 150 and 200 agents are shown in corresponding tables, where (1), (2) and (3) denote three intervals for calculating the collision rate.}

\subtable[50 Agents]{
\begin{tabular}{ |p{1.5cm}||p{0.8cm}|p{0.8cm}|p{0.8cm}|p{0.8cm}| }
 \hline

 Methods & (1) & (2) & (3) & avg \\
 \hline
 Y-net & 2.8\%  & 2.9\% & 3.8\% & 3.2\%\\ 
 \hline
 S-CSR & 2.5\%  & 1.7\% & 1.9\% & 2.0\%\\ 
 \hline 
 NSP(ours) & 0.6\%  & 0.6\% & 0.6\% & 0.6\% \\
 \hline 
\end{tabular}
}
\subtable[74 Agents]{
\begin{tabular}{ |p{1.5cm}||p{0.8cm}|p{0.8cm}|p{0.8cm}|p{0.8cm}| }
 \hline

 Methods & (1) & (2) & (3) & avg \\
 \hline
 Y-net & 3.8\%  & 4.8\% & 3.0\% & 3.9\%\\ 
 \hline
 S-CSR & 0.9\%  & 0.9\% & 1.5\% & 1.1\%\\ 
 \hline 
 NSP(ours) & 0.2\%  & 0.3\% & 0.5\% & 0.3\% \\
 \hline 
\end{tabular}
}
\subtable[100 Agents]{
\begin{tabular}{ |p{1.5cm}||p{0.8cm}|p{0.8cm}|p{0.8cm}|p{0.8cm}| }
 \hline
 Methods & (1) & (2) & (3) & avg \\
 \hline
 Y-net  & 4.2\% & 5.2\%  & 7.6\% & 5.7\%\\ 
 \hline
 S-CSR & 0.9\% & 1.1\%  & 0.8\% & 0.9\%\\ 
 \hline 
 NSP(ours) & 0.3\% & 0.7\%  & 0.4\% & 0.5\%\\
 \hline 
\end{tabular}
}
\subtable[150 Agents]{
\begin{tabular}{ |p{1.5cm}||p{0.8cm}|p{0.8cm}|p{0.8cm}|p{0.8cm}| }
 \hline
 Methods & (1) & (2) & (3) & avg \\
 \hline
 Y-net & 4.9\%  & 4.0\% & 3.4\% & 4.1\% \\ 
 \hline
 S-CSR & 0.6\%  & 1.0\% & 1.7\% & 1.1\%\\ 
 \hline 
 NSP(ours) & 0.2\%  & 0.5\% & 1.0\% & 0.6\%\\
 \hline
\end{tabular}
}
\subtable[200 Agents]{
\begin{tabular}{ |p{1.5cm}||p{0.8cm}|p{0.8cm}|p{0.8cm}|p{0.8cm}| }
 \hline
 Methods & (1) & (2) & (3) & avg \\
 \hline
  Y-net &  5.9\% & 4.0\%  & 3.5\% & 4.5\%\\ 
 \hline
 S-CSR & 0.6\% & 0.9\%  & 2.0\% & 1.2\%\\ 
 \hline 
 NSP(ours) & 0.2\% & 0.5\%  & 0.8\% & 0.5\%\\
 \hline
\end{tabular}
}
\label{tab:collision_supp}
\end{table}

\noindent
\textbf{\Large A Additional Experiments}
\vspace{15pt}

\noindent
\textbf{\large A.1 Generalization to Unseen Scenarios}
\vspace{10pt}

\noindent
We use the collision rate to evaluate prediction plausibility. We first elaborate on the definition of the collision rate and then show more experimental results. Provided there are N agents in a scene, we consider their collision rates during a period of time such as 4.8 seconds which is widely used to evaluate trajectory predictions~\cite{mangalam2020not,mangalam2021goals,zhou2021sliding}. We count one collision if the minimum distance between two agents is smaller than 2r at any time, where r is the radius of a disc representing an agent. The maximum possible number of collisions is $N(N-1)/2$. The final collision rate is defined as:
\begin{equation}
    R_{col} = \frac{M}{N(N-1)/2} 
\end{equation}
where $M$ is the number of collisions. 

We show more results on the scene, coupa0, with different numbers of agents. We chose coupa0 because it is a large space and can accommodate many people. The highest number of people simultaneously in the environment in the original data is merely 11. Therefore, this is a good scene to show how different methods can generalize to higher densities when learning from low density data. In each experiment, the agents are randomly initialized with different initial positions, initial velocities and goals near the boundary of the scene, which is sufficient for our method to simulate. Therefore, we use NSP to predict trajectories of 30 seconds (t = 0 to 29) at FPS = 10 for all agents. We sample three intervals out of every trajectory, from t = 0 to 8, t = 4 to 12 and t = 8 to 16, where the density in the central area reaches the highest during t=8 to 16. For each interval (8 seconds long), we subsample at FPS = 2.5 to get 20 frames, where the first 8 frames are used as input for Y-net~\cite{mangalam2021goals} and S-CSR~\cite{zhou2021sliding}. The remaining 12 frames and the predictions (12 frames) of Y-net and S-CSR are used to calculate the collision rate. Before prediction, all methods are trained on the training dataset of SDD under the same setting explained in the main paper.

\begin{figure}[tb]
\centering
\subfigure[Collision Rate]{\includegraphics[width=0.47\textwidth]{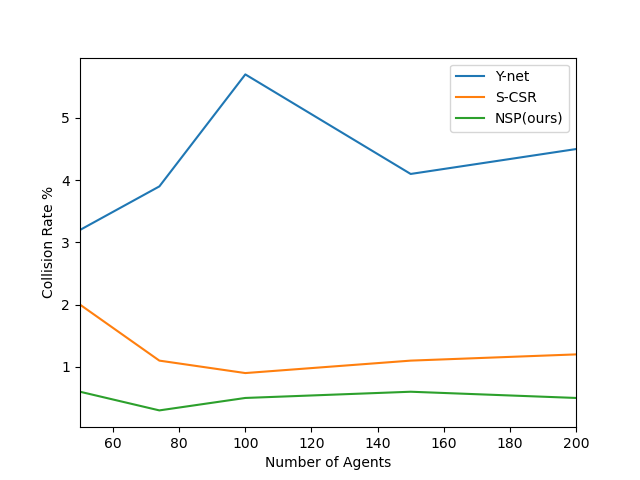}} 
\subfigure[Number of Collisions]{\includegraphics[width=0.47\textwidth]{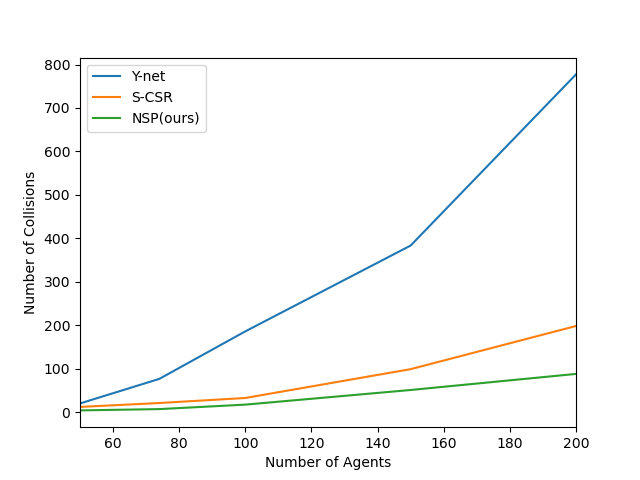}}
\caption{The collision rate and the number of collisions against the number of agents are shown in (a) and (b) respectively. Both of horizontal axes represent the number of agents from 50 to 200. The vertical axes in (a) and (b) represent the collision rate and the number of collisions respectively.  }
\label{fig:col_all_supp}
\end{figure}

\begin{figure}[tb]
\centering
\subfigure[74 Agents]{\includegraphics[width=0.47\textwidth]{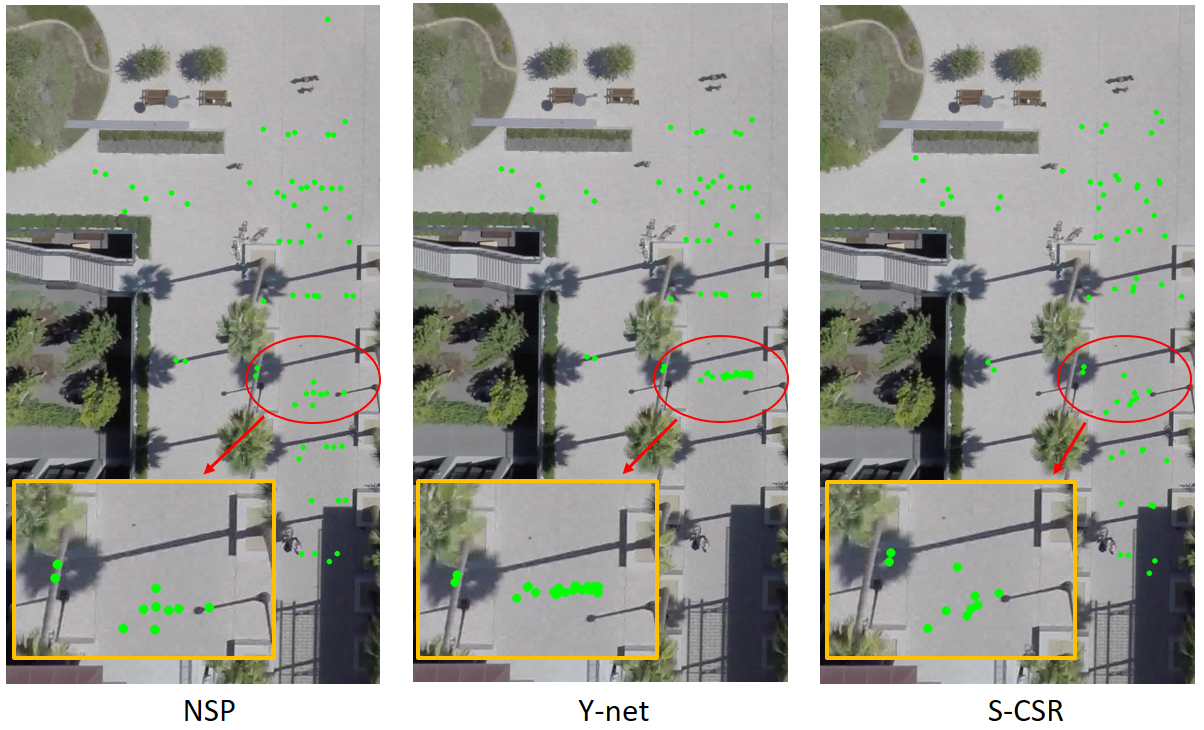}} 
\subfigure[100 Agents]{\includegraphics[width=0.47\textwidth]{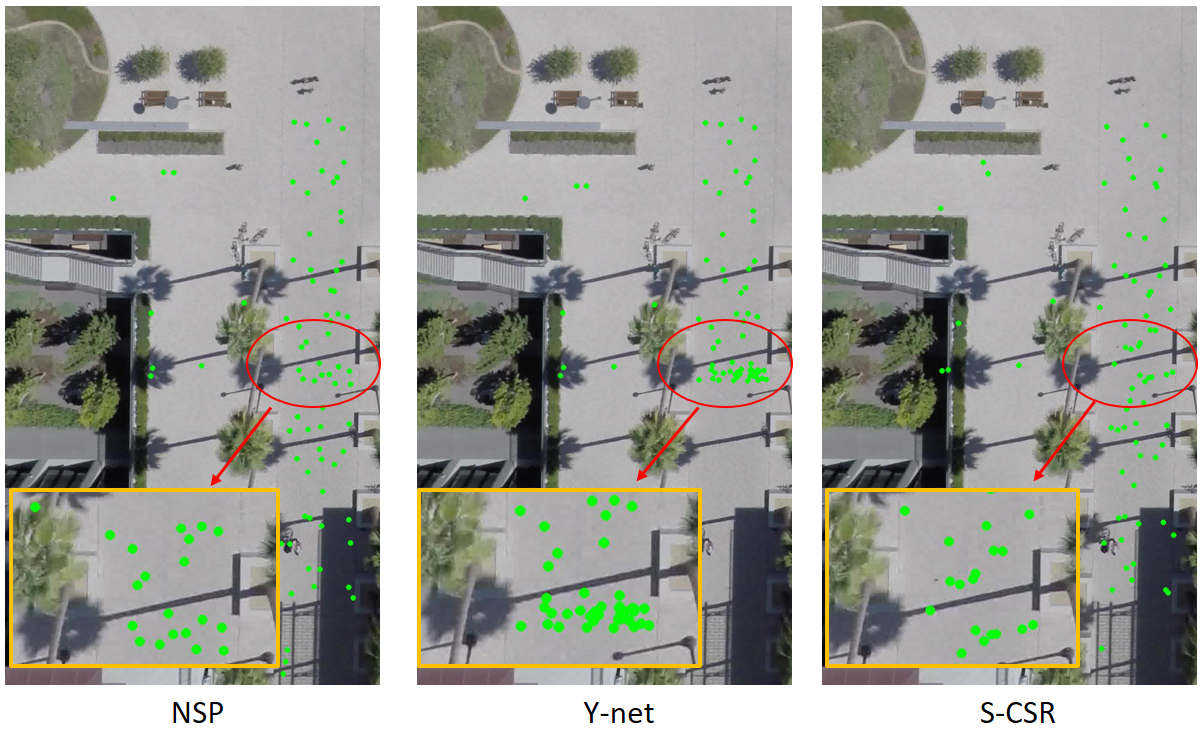}}
\\ 
\centering
\subfigure[150 Agents]{\includegraphics[width=0.47\textwidth]{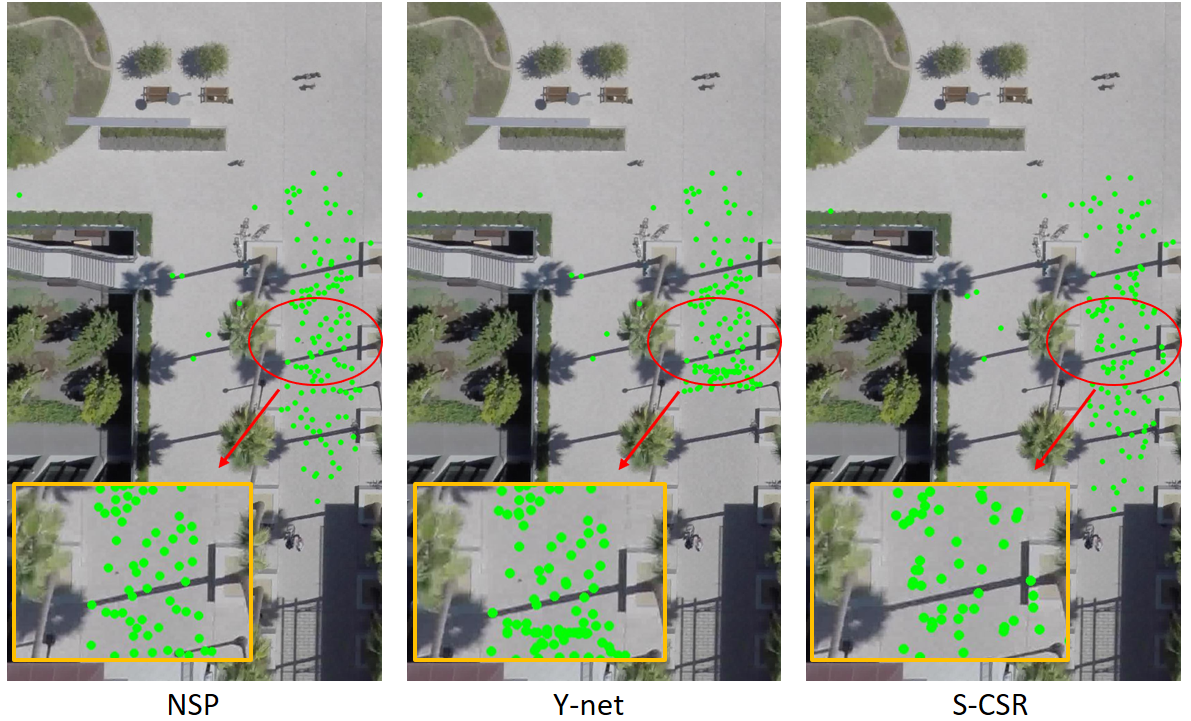}} 
\subfigure[200 Agents]{\includegraphics[width=0.47\textwidth]{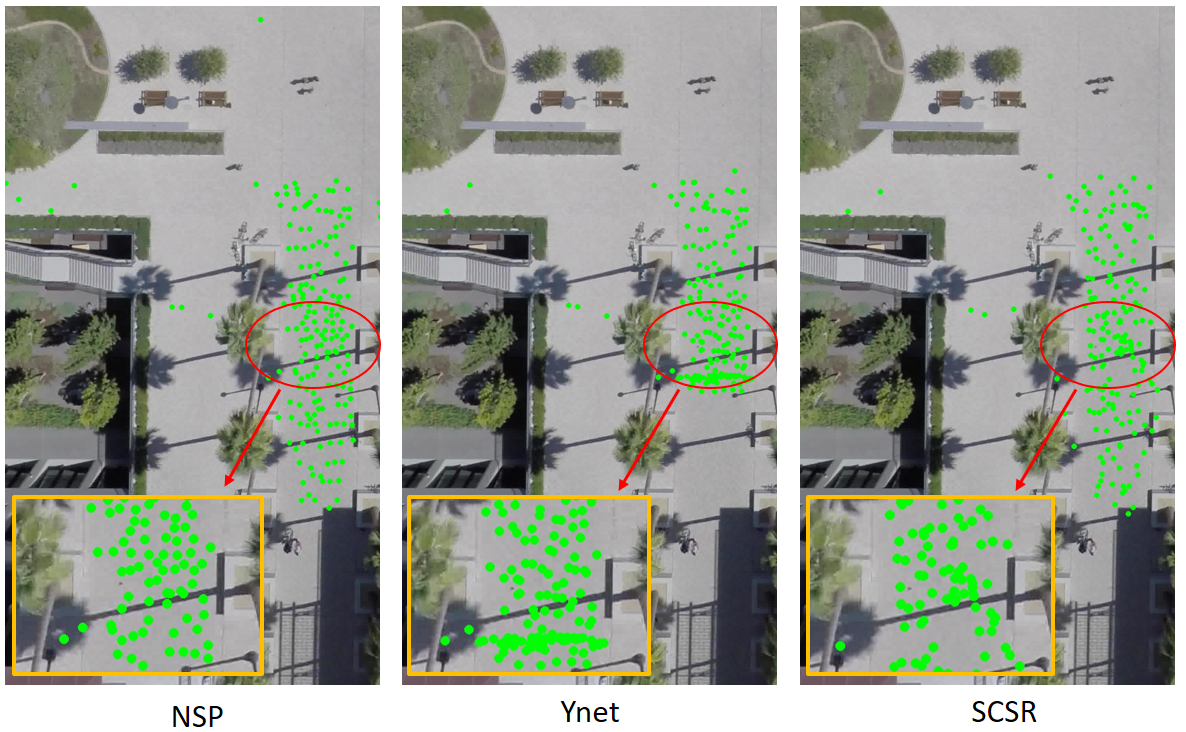}}
\caption{The visualization results of generalization to 74, 100, 150 and 200 agents on coupa0 are shown in (a), (b), (c) and (d) respectively. For each experimental setting, visualization results of NSP, Y-net and S-CSR are at the same frame. We amplify the area of red ellipse to boxes with yellow borders for better visualization performance.}
\label{fig:col_vis_supp}
\end{figure}

The results are shown in \Tabref{collision_supp}. We tested 50, 74, 100, 150 and 200 agents on the aforementioned three methods including ours. We can see that our method is always the best in the collision rate under different settings. Although its collision rate increases with the growth of the number of agents, our method is still the best compared with the baselines and our predictions are more plausible. In addition, we also plot the relation between the collision rate (and the number of collisions) and the agent number ranging from 50 to 200 in \Figref{col_all_supp}. Y-net is worse than S-CSR and NSP. In addition, although the trend of NSP and S-CSR are similar, the number of collisions of S-CSR increases faster than NSP. Finally, some visualization results can be found in~\Figref{col_vis_supp}. Here, every green disc has a radius of 7.5 pixels. When two green discs intersect, they collide with each other.~\Figref{col_vis_supp} demonstrates that our method (NSP) has better performance in avoiding collisions than Y-net and S-CSR.

\vspace{15pt}
\noindent
\textbf{\large A.2 Interpretability of Prediction}
\vspace{10pt}

\begin{figure}[tb]
\centering
\includegraphics[width=\textwidth]{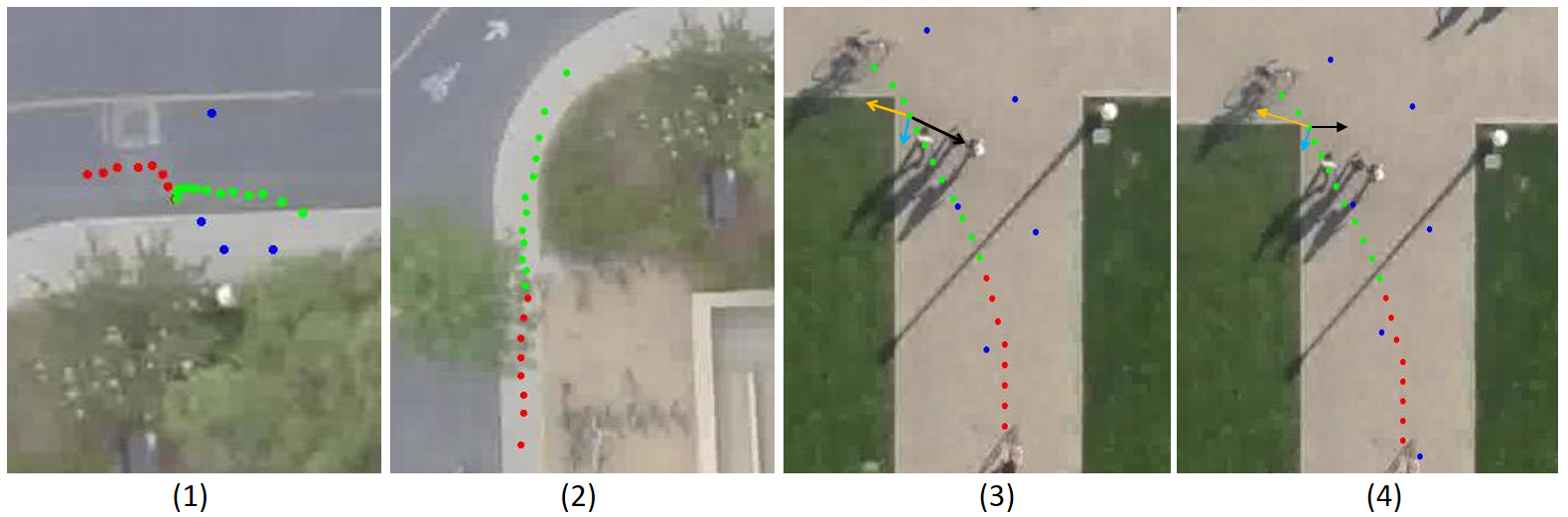}
\caption{Examples of interpretability. Red dots are observed, green dots are our prediction. Bule dots in (1), (3) and (4) are other pedestrians at time step 7, 16 and 17 respectively. We show the influence of all forces, $F_{goal}$, $F_{col}$ and $F_{env}$, on the whole trajecroty in (1) and (2). We display detailed analysis of three forces at two consecutive time steps of the same agent, where $F_{goal}$, $F_{col}$ and $F_{env}$ are shown as yellow, light blue and black arrows respectively. }
\label{fig:interpret_supp}
\end{figure}

\begin{figure}[tb]
\centering
\includegraphics[width=0.8\textwidth]{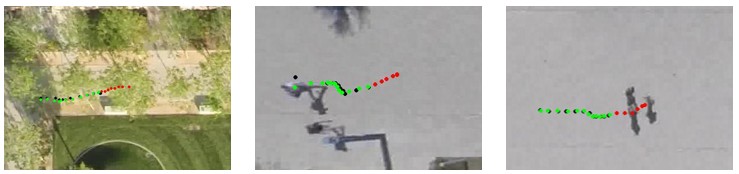}
\caption{Motion randomness is captured by our model. Red dots are observed, green dots are our prediction and black dots are the
ground-truth.}
\label{fig:randomness_supp}
\end{figure}

\noindent
More examples of interpretability are shown in~\Figref{interpret_supp}. In \Figref{interpret_supp} (1)-(2), we show the influence of different three forces, $F_{goal}$, $F_{col}$ and $F_{env}$, on the whole trajectory of an agent. In \Figref{interpret_supp} (3)-(4), we choose two consecutive moments of one agent for analysis. In \Figref{interpret_supp} (1), instead of directly aiming for the goal, the agent suddenly turns (at the intersection between red and green dots) due to the incoming agents (the three blue dots under the green dots). The result is a result of major influence from $F_{goal}$ and $F_{col}$. Similarly, the agent in \Figref{interpret_supp} (2) did not need to avoid other agents but still did not directly walk towards the goal, because of $F_{env}$ from the grass. In \Figref{interpret_supp} (3)-(4), we show the detailed analysis of forces at two consecutive time steps of the same agent, where $F_{env}$ is from the lawn which is a 'weakly repulsive area'.  
More examples where randomness is captured by our model are shown in \Figref{randomness_supp}. 

\vspace{15pt}
\noindent
\textbf{\large A.3 Ablation Experiments}
\vspace{10pt}

\noindent
We conduct more ablation experiments to further validate our design decisions and explore the effect of components of our model. The ablation studies on the network architectures focuses on the Goal-Network and the Collision-Network. The main variants are with/without LSTM to show the importance of the temporal modeling for learning $\tau$ and $k_{nj}$, and replacing the MLPs with simple two-layer MLPs. \Tabref{ae1_supp} shows the results on SDD. We can see that the temporal modeling and the original MLPs make our model achieve the best performance.
To understand the role of each component in our model, we take social force model (SFM) as the baseline and incrementally add components from our model. The results are shown in \Tabref{ae2_supp}. We tried our best to manually find good parameter values: $\tau=0.5$, $k_{nj}$=25/50 and $k_{env}$=65. We adopted the same way with our model to sample destinations for SFM. Then we only learn $\tau$ and $k_{nj}$. At last, the result of the full model without CVAE is given. The performance is better when more components are added.

\begin{table}[t]
\begin{small}
\begin{center}
\caption{Ablation experiments on network architecture. Goal-Network and the Collision-Network possess the same architecture under each experimental setup.}
\label{tab:ae1_supp}
\begin{tabular}{ |p{1.8cm}||p{2.4cm}|p{1.8cm}|}
\hline
ADE & Two layers MLP & Full MLP \\
\hline
w/o LSTM & 6.83 & 6.61 \\ 
\hline
with LSTM & 6.66 & \textbf{6.52}\\
\hline
\end{tabular}
\end{center}
\end{small}
\end{table}

\begin{table}
\begin{small}
\begin{center}
\caption{Ablation experiments on SDD. Different components from our model are added incrementally}
\label{tab:ae2_supp}
\begin{tabular}{ |p{1.2cm}||p{2.0cm}|p{2.7cm}|p{1.1cm}|}
\hline
$k_{nj}$=25 & hand-tuned & learned $\tau$ and $k_{nj}$ & NSP\\
\hline
ADE & 8.32 & 6.53 & 6.52 \\ 
\hline
FDE & 10.97 & 10.61 & 10.61\\
\hline
\hline
$k_{nj}$=50 & hand-tuned & learned $\tau$ and $k_{nj}$ & NSP\\
\hline
ADE & 7.54 & 6.53 & 6.52 \\ 
\hline
FDE & 10.81 & 10.61 & 10.61\\
\hline
\end{tabular}
\end{center}
\end{small}
\end{table}

\vspace{15pt}
\noindent
\textbf{\Large B Details of the Neural Social Physics Model}
\vspace{15pt}

\noindent
In this section, we elaborate the details of the Goal Sampling Network (GSN) and the conditional Variational Autoencoder (CVAE) in our model.

\vspace{15pt}
\noindent
\textbf{\large B.1 Goal Sampling Network}
\vspace{10pt}

\begin{figure}[tb]
\centering
\includegraphics[width=0.8\textwidth]{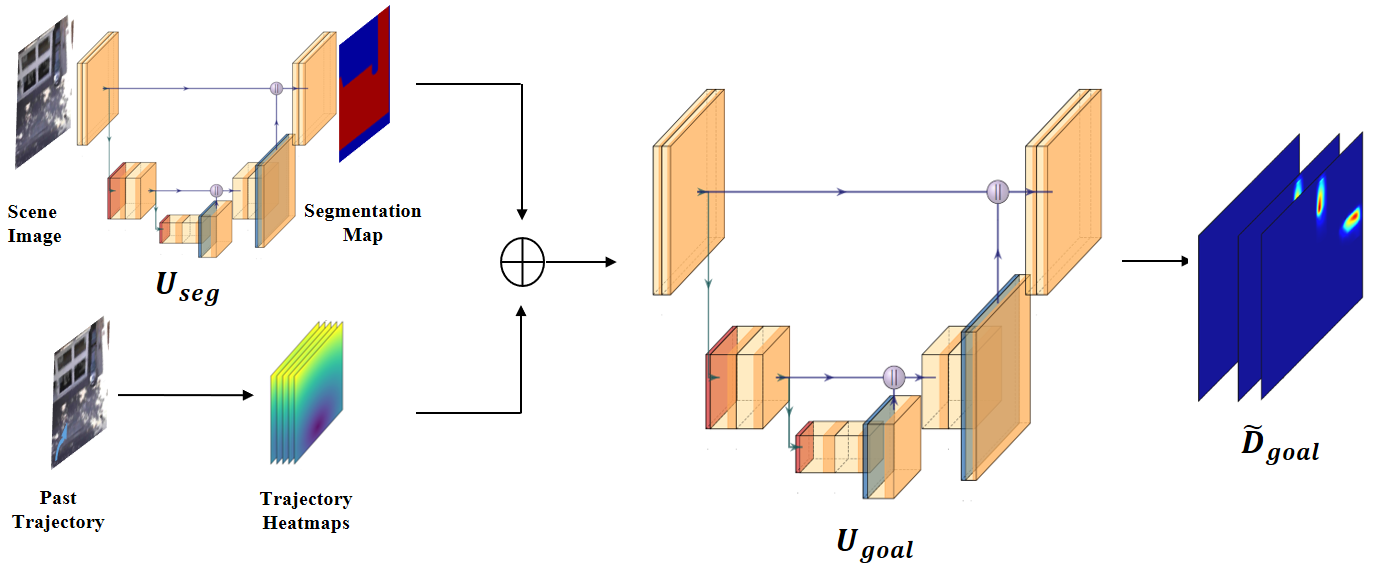}
\caption{Model Architecture of Goal Sampling Network. The detailed network architecture of two U-nets, $U_{seg}$ and $U_{goal}$, can be found in~\cite{mangalam2021goals}.}
\label{fig:gsn_supp}
\end{figure}

\noindent
The main components of the GSN are two U-nets~\cite{ronneberger2015u} as illustrated in~\Figref{gsn_supp}. We first feed the scene image $\textit{\textbf{I}}$ to a U-net, $U_{seg}$, to get its corresponding environment pixel-wise segmentation with dimension of $H*W*K_c$. $H$ and $W$ are the height and width of $\textit{\textbf{I}}$, and $K_c$ is the number of classes for segmentation. The segmantation maps are byproducts of the GSN from~\cite{mangalam2021goals}. NSP can use manually annotated or automatically segmented environment maps to calculate $F_{env}$, but using segmentation maps from the GSN is more efficient. Then the past trajectories $\{p^t\}_{t=0}^M$ are converted into M+1 trajectory heatmaps by:
\begin{equation}
    Hm(t,i,j) = 2\frac{\lVert(i,j) - p^t\rVert}{   \max \limits_{(x,y) \in \textit{\textbf{I}}} \lVert(x,y) - p^t\rVert}
\end{equation}
where $(i,j)$ is the pixel coordinate on the heatmap and $(x,y)$ is the pixel coordinate on the scene image $\textit{\textbf{I}}$. Then, we concatenate these trajectory heatmaps and the segmentation map to get the input with dimension of $H*W*(K_c+M+1)$ for the network $U_{goal}$. $U_{goal}$ will output a non-parametric probability distribution map, $\Tilde{D}_{goal}$, with dimensions $H*W$. Every pixel in $\Tilde{D}_{goal}$ has a corresponding probability value between 0 and 1, and their sum is equal to 1. Details of these two U-nets can be found in~\cite{mangalam2021goals}. We train the GSN by minimizing the Kullback–Leibler divergence between predicted $\Tilde{D}_{goal}$ and its ground truth $D_{goal}$. We assume that $D_{goal}$ is a discrete gaussian distribution with a mean at the position of the ground-truth goal and a hyper-parameter variance $\sigma_{goal}$. During testing, instead of picking the position with highest probability, we adopt the test-time sampling trick introduced by~\cite{mangalam2021goals} to sample goals for better performance.

\vspace{15pt}
\noindent
\textbf{\large B.2 Conditional Variational Autoencoder}
\vspace{10pt}

\begin{figure}[tb]
\centering
\includegraphics[width=\textwidth]{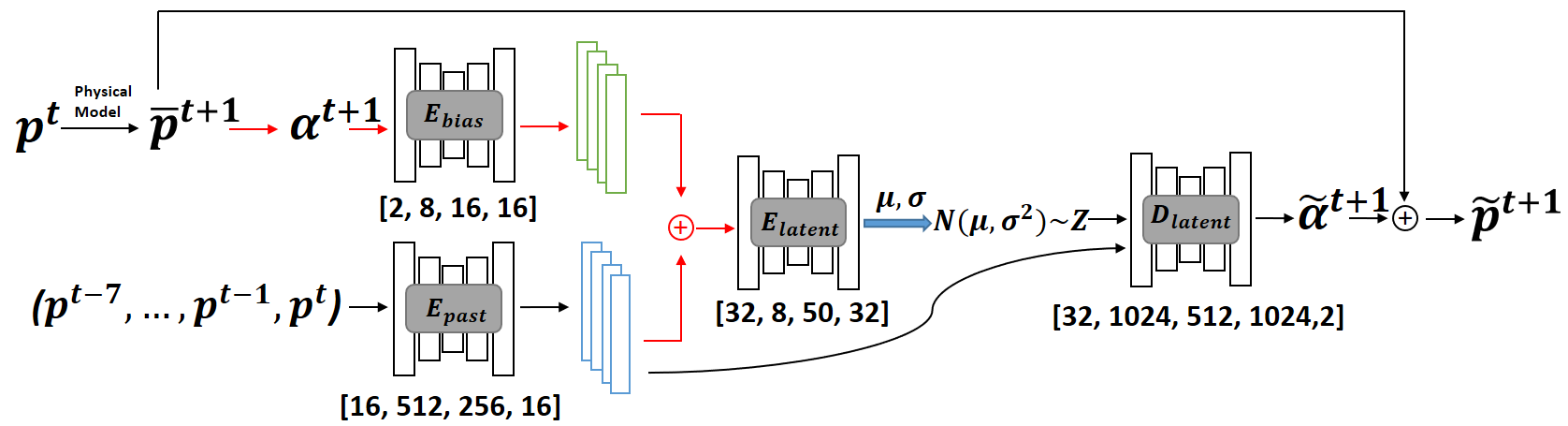}
\caption{The architecture of the CVAE, where $\Bar{p}^{t+1}$ is the intermediate prediction out of our force model and $\alpha^{t+1} = p^{t+1} - \Bar{p}^{t+1}$. Encoder $E_{bias}$, $E_{past}$, $E_{latent}$ and decoder $D_{latent}$ are all MLP networks with dimensions indicated in the square brackets. Red connections are only used in the training phase.}
\label{fig:cvae_supp}
\end{figure}

\noindent
We model the dynamics stochasticity for each agent individually by using a CVAE as illustrated in~\Figref{cvae_supp}. Red connections are only used in the training phase. Given an agent $p^t$ and his/her destination, a deterministic prediction $\bar{p}^{t+1}$ without dynamics stochasticity is first calculated from $F_{goal}$, $F_{col}$ and $F_{env}$ and a semi-implicit scheme. During training time, we use the corresponding ground truth $p^{t+1}$ to calculate the error $\alpha^{t+1} = p^{t+1} -  \bar{p}^{t+1}$, and feed $\alpha^{t+1}$ into an encoder $E_{bias}$ to get the feature $f_{bias}$. The brief history $(p^{t-7}, \dots, p^{t-1}, p^t)$ is encoded as $f_{past}$ by using an encoder $E_{past}$. We concatenate $f_{bias}$ with $f_{past}$ and encode it using a latent encoder to yield the parameters $(\mu, \sigma)$ of the gaussian distribution of the latent variable Z. We sample Z, concatenate it with $f_{past}$ for history information, and decode using the decoder $D_{latent}$ to acquire our guess for stochasticity $\Tilde{\alpha}^{t+1}$. Finally, the estimated stochasticity will be added to the deterministic prediction $\bar{p}^{t+1}$ to get our final prediction $\Tilde{p}^{t+1}$. During testing time, the ground truth $p^{t+1}$ is unavailable. Therefore, we sample the latent variable Z from a gaussian distribution $N(0, \sigma_{latent}I)$ where $\sigma_{latent}$ is a hyper-parameter. We concatenate the sampled Z and $f_{past}$ to decode directly using the learned decoder $D_{latent}$ to get the estimate of stochasticity $\Tilde{\alpha}^{t+1}$. We can produce final prediction $\Tilde{p}^{t+1}$ using the same way as the training phase. Encoders $E_{bias}$, $E_{past}$, $E_{latent}$ and the decoder $D_{latent}$ are all multi-layer perceptrons (MLP) with dimensions indicated in the square brackets in~\Figref{cvae_supp}.

\begin{table}[t]
\begin{center}
\caption{Hyper-parameters for all six datasets.}
\begin{tabular}{ |p{1.8cm}||p{1.3cm}|p{1.3cm}|p{1.3cm}|p{1.3cm}|p{1.3cm}|p{1.3cm}|}
 \hline
 Hyper-Para & ETH & Hotel & UNIV & ZARA1 & ZARA2 & SDD \\
 \hline
 a($\tau$) & 1 & 1 & 1 & 1 & 1 & 1 \\
 \hline
 b($\tau$) & 0.1 & 0.1 & 2.2 & 1.6 & 1.4 & 0.4 \\
 \hline
 a($k_{nj}$) & 50 & 50 & 50 & 50 & 50 & 100 \\
 \hline
 b($k_{nj}$) & 0 & 0 & 0 & 0 & 0 & 0 \\
 \hline
 $\omega$ & $\pi/3$ & $\pi/3$ & $\pi/3$ & $\pi/3$ & $\pi/3$ & $\pi/3$\\
 \hline
 $r_{col}$ & 75 & 75 & 75 & 75 & 75 & 100 \\
 \hline
 $r_{env}$ & 50 & 50 & 50 & 75 & 75 & 50 \\
 \hline
 $\sigma_{goal}$ & 4 & 4 & 4 & 4 & 4 & 4 \\
 \hline
 $\sigma_{latent}$ & 1.3 & 1.3 & 1.3 & 1.3 & 1.3 & 1.3 \\
 \hline 
 $\lambda_{weak}$ & N/A & N/A & N/A & N/A & N/A & 0.2 \\
 \hline
 
\end{tabular}

\label{tab:hyperp_supp}
\end{center}
\end{table}

\vspace{15pt}
\noindent
\textbf{\large C Implementation Details}
\vspace{15pt}

\noindent
We use ADAM as the optimizer to train the Goal-Network, Collision-Network and $F_{env}$ with a learning rate between $3 \times 10^{-5}$ and $3 \times 10^{-4}$, and to train the CVAE with a learning rate between $3 \times 10^{-6}$ and $3 \times 10^{-5}$. When we train the CVAE of our model, the training data is scaled by 0.005 to balance reconstruction error and KL-divergence in $l_{cvae}$. The hyper-parameter $\lambda$ in $l_{cvae}$ is set to 1. Concrete structures of all sub-network are shown in \Figref{cvae_supp}. In all experiments, we follow existing work in using 20-frame samples. For trajectories with less than 20 frames, we treat them as observed dynamic obstacles and do not predict their trajectories. 

For the Goal-Network, instead of learning parameter $\tau$ directly, we set $\tau = a*sigmoind(NN_{\phi_1}(q^t, p^T))+b$ where $a$ and $b$ are hyper-parameters. We list all hyper-parameters of our model in \Tabref{hyperp_supp}. We segment scene images into two classes and three classes on ETH/UCY and SDD, respectively. The two classes on ETH/UCY are `walkable area' and `unwalkable area'. Three classes on SDD include `walkable area', `unwalkable area' and `weakly repulsive area' that some people tend to avoid such as lawns. The calculation of $F_{env}$ on ETH/UCY has been introduced in our main paper. On SDD, we calculate the position of the obstacle $p_{obs}$ and the position of the weak obstacle $p_{w-obs}$ (i.e. in the weakly repulsive area) by averaging pixels that are classified as `unwalkable area' and `weak repulsive area' respectively. Then, the $F_{env}$ consists of two repulsive forces from  $p_{obs}$ and $p_{w-obs}$ as shown in \Eqnref{env_supp}, where the parameter $k_{env}$ is shared and an additional hyper-parameter $\lambda_{weak}$ is introduced for $p_{w-obs}$:
\begin{equation}
\label{eq:env_supp}
    F_{env} = \frac{k_{env}}{\lVert p_n^t - p_{obs}\rVert} (\frac{p_n^t - p_{obs}}{\lVert p_n^t - p_{obs}\rVert}) + \frac{\lambda_{weak}k_{env}}{\lVert p_n^t - p_{w-obs}\rVert} (\frac{p_n^t - p_{w-obs}}{\lVert p_n^t - p_{w-obs}\rVert}) 
\end{equation}

\end{document}